\documentclass[pdflatex,sn-mathphys-num,pagebackref,breaklinks,colorlinks]{sn-jnl}
\usepackage{enumitem}

\usepackage{graphicx}%
\usepackage{multirow}%
\usepackage{amsmath,amssymb,amsfonts}%
\usepackage{amsthm}%
\usepackage{mathrsfs}%
\usepackage[title]{appendix}%
\usepackage[table]{xcolor}
\usepackage{booktabs}
\usepackage{xcolor}%
\usepackage{textcomp}%
\usepackage{manyfoot}%
\usepackage{soul}
\usepackage{booktabs}%
\usepackage{algorithm}%
\usepackage{algorithmicx}%
\usepackage{algpseudocode}%
\usepackage{listings}%
\usepackage{comment}

\theoremstyle{thmstyleone}%
\theoremstyle{thmstyletwo}%

\theoremstyle{thmstylethree}%

\begin{document}

\title{RBF Weighted Hyper-Involution for RGB-D Object Detection}


\author[1]{\fnm{Mehfuz A.} \sur{Rahman}}\email{mehfuza.rahman@smu.ca}

\author[2]{\fnm{Khushal} \sur{Das}}\email{khushal.das@dimes.unical.it}

\author*[1]{\fnm{Jiju} \sur{Poovvancheri}}\email{jiju.poovvancheri@smu.ca}

\author[3]{\fnm{Neil} \sur{London}}\email{nlondon@modesttree.com}

\author[4]{\fnm{Dong} \sur{Chen}}\email{chendong@njfu.edu.cn}

\affil*[1]{\orgdiv{Graphics and Spatial Computing Lab}, \orgname{Saint Mary's University}, \orgaddress{\city{Halifax}, \postcode{B3H3C3}, \state{Nova Scotia}, \country{Canada}}}

\affil[2]{\orgdiv{DIMES}, \orgname{University of Calabria}, \orgaddress{ \city{Rende}, \postcode{87036}, \state{CS}, \country{Italy}}}

\affil[3]{\orgname{Modest Tree Media Inc.}, \orgaddress{\city{Halifax}, \postcode{B3H3C3}, \state{Nova Scotia}, \country{Canada}}}

\affil[4]{\orgname{Nanjing Forestry University}, \orgaddress{ \city{Nanjing}, \country{China}}}


\abstract{A vast majority of augmented reality devices come equipped with depth and color cameras. Despite their advantages, extracting both photometric and depth features simultaneously in real-time remains challenging due to inherent differences between depth and color images. Furthermore, standard convolution operations are insufficient for extracting information directly from raw depth images, leading to inefficient intermediate representations. To address these issues, we propose a real-time two-stream RGBD object detection model. Our model introduces two new components: a dynamic radial basis function (RBF) weighted depth-based hyper-involution that adjusts dynamically based on spatial interaction patterns in raw depth maps, and an up-sampling based trainable fusion layer that combines extracted depth and color image features without obstructing information transfer between them. Experimental results demonstrate that the proposed approach achieves the strongest performance among existing RGB-D 2D object detection methods on NYU Depth V2, while remaining competitive on the SUN RGB-D benchmark.}




\maketitle

\section{Introduction}
\label{sec:intro}

Object detection is essential in autonomous robotics and AR applications, focusing on identifying and locating objects within a scene. Research~\cite{gupta2014learning,xiao2021fetnet, redmon2016you} shows that RGB-D based detection outperforms RGB-based methods. Depth images complement RGB detection in several ways: they enhance visibility of object boundaries, aiding in precise location and coverage, especially under poor lighting or shadows; they address scale distortions from perspective projections, helping detectors learn object sizes; they can detect hidden objects blending into backgrounds; and they handle misleading colors and textures that could mislead classification.

\begin{figure}[h]
	\centering
	\includegraphics[width=0.9\linewidth]{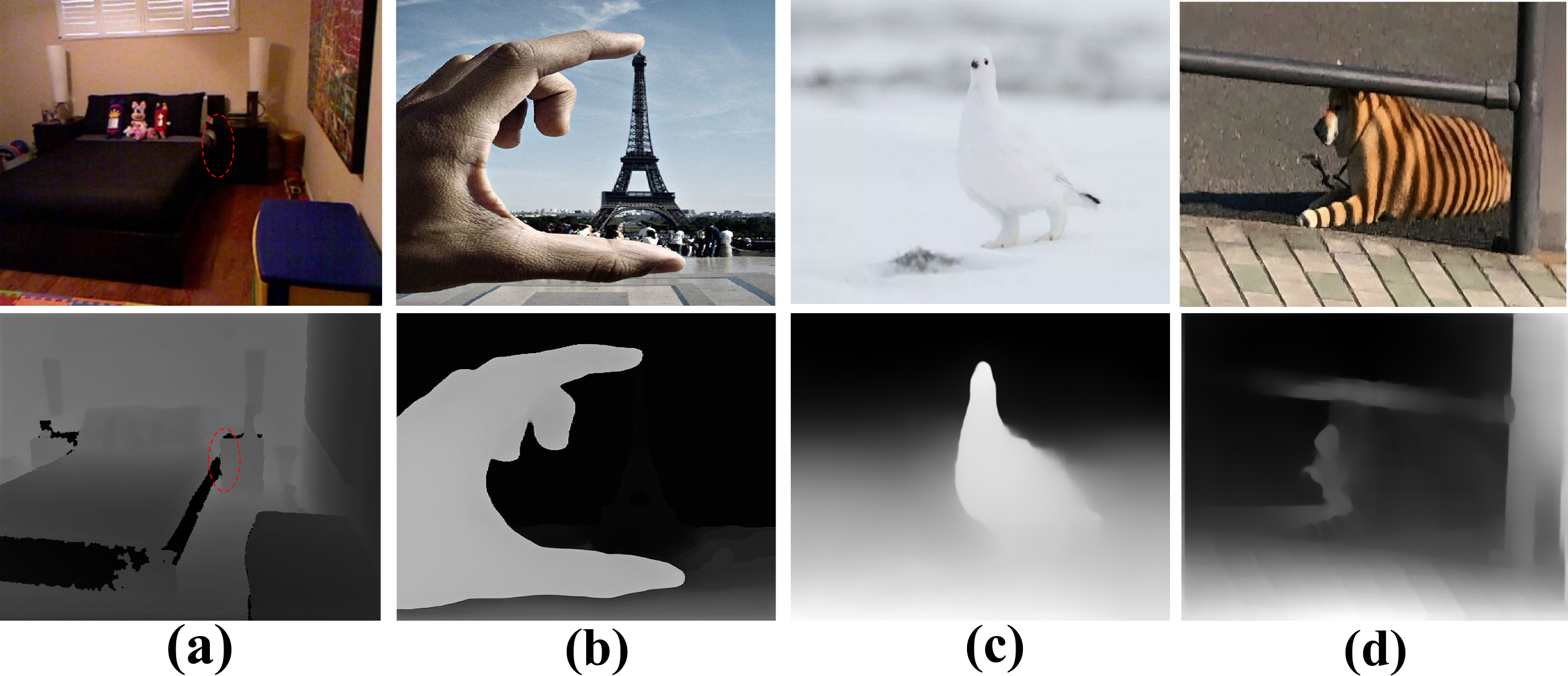}
	\caption{Few instances where the usefulness of depth for object detection is visible. Image courtesy: \cite{Silberman:ECCV12, polseno_2020, ranftl2021vision, rankuzz.com_2020, fun_dog_tiger}}
	\label{DepthBenefit}
\end{figure}

The rapid advancement and growing accessibility of affordable commercial depth sensors over the past decade have significantly contributed to the adoption of RGB-D based detection in both research and real-world applications. Depth sensors are now an integral component of many modern AR headsets (e.g., Microsoft HoloLens 2), enabling them to capture depth images, or depth maps, where each pixel encodes the distance of a specific point in the scene from the sensor. When combined with their corresponding color images, these depth images form four-channel RGB-D (red, green, blue, depth) data, enhancing object detection capabilities. Prior state-of-the-art research~\cite{gupta2014learning,xiao2021fetnet} has already proven the significance and  performance improvement of RGB-D based object detection over RGB based detection. Depth images complement RGB based object detection in multiple ways. Firstly, depth images better visualize object boundaries, making it easier to locate objects and properly cover them with bounding boxes. This is particularly important in cases where the object boundaries are not clear in color images due to poor illumination or heavy shadows, as shown in Figure \ref{DepthBenefit}(a). Secondly, depth images can resolve scale distortions that often appear in color images due to perspective projections. Depth images provide useful information to object detectors, making it easier to learn the relative sizes of objects in a scene. One such phenomenon is illustrated in Figure \ref{DepthBenefit}(b). Thirdly, depth images can detect camouflaged objects that might not be easily visible in color images due to their similarity to their background which is demonstrated with the RGB and corresponding depth map of a penguin in Figure \ref{DepthBenefit}(c). Finally, depth images can handle delusive color and texture in images (Figure \ref{DepthBenefit}(d)) that can mislead the object classification if solely relied on color and texture information.

Despite the clear benefits of incorporating depth information, processing depth maps alongside color images in object detection models is difficult due to their fundamental differences. Consequently, recent RGB-D object detection methods use two-stream networks, which separately extract features from color and depth images before merging them at specific points in the model \cite{ gupta2014learning, ophoff2018improving, ophoff2019exploring}.

Most fusion stages for depth and color features are naively chosen, often relying on simple concatenation without proper learnable parameters for effective backpropagation, hindering effective information exchange \cite{xiao2021fetnet}. Some researchers encode depth maps into different representations, which is time-consuming and based on intuition \cite{gupta2014learning,li2018cross,xu2017multi}. The standard convolutional operations are designed for color images, not raw depth images. An alternative to standard convolution is needed for processing raw depth images. Many existing RGB-D models use slow two-stage detectors from the outdated RCNN series, unlike faster modern real-time models \cite{ tan2020efficientdet, wang2021scaled}. 

We tackle these challenges with a depth-aware involution-based fusion network for RGB-D object detection. By using RBF weighting of pixel depths during feature extraction, our model effectively captures depth information. Our real-time, single-stage architecture (Figure \ref{MainModel}) introduces two high-performance components. The key contributions are:

\begin{itemize}[]
    
\item A dynamic depth aware hyper-involution module as an alternative to standard convolution for proper utilization of raw depth information and spatial specific features.
\item An improved encoder-decoder fusion stage in the middle layers can combine features from the depth and RGB streams, extracting the most significant semantic information.
\item We introduce a novel outdoor RGB-D dataset for 2D object detection, with carefully curated annotations. A thorough characterization of the dataset, including statistics, annotation methodology, depth synthesis procedures, and qualitative samples, is provided in Section\ref{sec_novel_data}.
\end{itemize}

\section{Related Work}

In recent years, the computer vision community has fervently introduced a plethora of state-of-the-art models for conventional RGB-based object detection. The object detection architectures can be categorized into two groups namely: single stage and two-stage detectors \cite{zaidi2022survey}. Single stage detectors predict the position and class label of the object within an image in a single pass through the neural network without the need for additional region proposals or refining components. At the moment, the leading single stage models \cite{wang2022yolov7,wang2021scaled, wang2021you} are the successors of YOLO \cite{redmon2016you,redmon2017yolo9000, redmon2018yolov3, yolov8, yolo11, yolo12} and FCOS \cite{tian2019fcos, tian2020fcos} series. Conversely, two-stage detectors use a combination of two neural networks to detect objects in the image. First, the region proposal network (RPN) generates a set number of potential locations where objects may be present in the image. These proposals are then passed to the detection network which refines the location and identification of the objects in the proposals. Some latest additions to state-of-the-art two-stage models include \cite{hong2022dynamic,sun2021sparse}. Overall, these RGB based detection models mainly introduce various components in their extended architecture to compete for speed and accuracy ignoring the importance of cross modal perception. While traditional 2D object detection methods remain the primary focus of this work, it is worth noting that they have laid the groundwork for subsequent advances in 3D detection. In particular, significant progress has been achieved in 3D object detection using LiDAR data \cite{fan2023once,li2022deepfusion,fan2023once} as well as RGB-D representations \cite{li2024cp,wang2023octformer}. A comprehensive overview of recent developments in point cloud learning is provided by Guo et al. \cite{9127813}, which outlines current research trends and emerging directions in the field.

In this paper, we investigate the research challenges of RGB-D based object detection and develop an improved model for RGB-D based object detection. It should be noted that closely related problems such as salient object detection (SOD) \cite{9810116, 10179145} and object recognition \cite{10124000} from RGB-D images are gaining a lot of attention recently. While SOD aims at highlighting visually salient object regions, object recognition aims to classify single object images. Thus the algorithmic designed results are different from RGB-D object identification and localization. Therefore, we focus on various existing RGB-D object detection architectures including their limitations followed by brief studies on alternatives to standard convolution and hyper-networks which are core components of our RGB-D based detection. 

\subsection{Detection based on HHA Formats}

Gupta et al. \cite{gupta2014learning} introduced a fusion-based model for RGB-D object detection and a geocentric embedding technique converting raw depth images to the HHA format (Horizontal disparity, Height above ground, Angle concerning gravity direction). This conversion process, however, is time-consuming \cite{hazirbas2016fusenet}. Later, \cite{gupta2016cross} tackled the lack of depth data by using supervision transfer from a pre-trained RGB backbone, improving accuracy but relying on a non-real-time two-stage Fast RCNN detector \cite{girshick2015fast}. Xu et al. \cite{xu2017multi} built on this by proposing a three-stream model with supervision transfer, but it also used the HHA conversion 
and had high computational costs due to three parallel backbones and separate Faster RCNN heads \cite{krizhevsky2012imagenet}. Chen et al. \cite{chen2023fafnet} introduced FAFNet, excelling on benchmarks but struggling with small object segmentation. Wang et al. \cite{wang2024amnet} presented AMNet, enhancing fusion accuracy through attention and multi-modality but faced potential information loss and weak inter-modality correlation. Li et al. \cite{li2018cross} proposed the Cross-Modal Attentional Context (CMAC) algorithm using LSTM \cite{hochreiter1997long} and Spatial Transformer Networks (STN) \cite{jaderberg2015spatial} for object part identification, but it suffered from high memory use, overfitting, and sensitivity to random weight initialization due to its reliance on HHA conversion.

\subsection{Detection using Raw Depth Maps}

Recent RGB-D object detection now uses raw depth maps, avoiding HHA conversion but facing some challenges. Zhang et al. \cite{zhang2020two} created a three-stream model with Channel Weights Fusion (CWF), limited by its focus on indoor human patterns. Ophoff et al. \cite{ophoff2018improving} used a single-stage detector for real-time RGB-D pedestrian detection, but it suffers from simple concatenation and high feature dimensions. Their work \cite{ophoff2019exploring} added multi-class detection with dimension-reducing convolution, needing separate pre-training for depth and RGB. Xiao et al. \cite{xiao2021fetnet} improved feature flow between depth and RGB but lacked optimization in depth data extraction. Bi et al. \cite{bi2022moving} introduced a framework for edge-guided depth map super-resolution and moving object detection, but fixed hyperparameters could cause suboptimal results. More recently, transformer-based and cross-modal architectures have explored depth-guided feature interaction beyond traditional convolutional operators, particularly in RGB-D settings, demonstrating the continued relevance of adaptive spatial filtering mechanisms \cite{pan2024multi}. Further, recent multimodal studies emphasize the importance of depth-driven feature interaction, showing that explicit modeling of depth relationships can improve robustness in complex scenes \cite{depth2024dipformer}.  The transformer-based RGB-D approaches also exhibit limitations. Pan et al. \cite{pan2024multi} rely on computationally expensive transformer backbones and auxiliary depth-completion tasks, which limit real-time applicability, while depth-interaction transformer models such as DiPFormer \cite{depth2024dipformer} are primarily designed for semantic segmentation and incur high memory overhead, making them less suitable for lightweight RGB-D object detection.

\subsection{Alternatives to Standard Convolution}

In recent years, various flexible and effective alternatives to the standard convolution operation \cite{lecun1998gradient} have emerged. Some adapt dynamically using pixel information, while others use depth. For instance, deformable convolution \cite{dai2017deformable} learns geometric transformations such as scale, pose, and part deformation. A faster, lighter version, Deformable ConvNets v2 (DCNv2) \cite{zhu2019deformable}, was introduced to address issues with irrelevant image regions. Pixel Adaptive Convolution (PAC) \cite{su2019pixel} adapts to image content while maintaining the benefits of standard convolution. Conditionally parameterized convolution, or CondConv \cite{yang2019condconv}, learns based on specific input samples, while dynamic convolution \cite{chen2020dynamic} adapts by superimposing multiple kernels.

Several studies have utilized depth maps to manipulate convolution kernels. Depth-aware convolution and average pooling \cite{wang2018depth} focus on pixels with similar depth values. S-conv \cite{chen2021spatial} improves segmentation by applying dimensional information to filter weights, generating location-adaptive filters. ShapeConv \cite{cao2021shapeconv} uses depth maps to extract content and location information, enhancing semantic segmentation accuracy. 
Depth-wise convolution \cite{tan2019efficientnet} aims to improve neural network efficiency but should not be confused with depth-based convolution.
Each alternative has limitations. DCNv2 \cite{zhu2019deformable} is slower and more parameter-intensive than standard convolution. CondConv \cite{yang2019condconv} and dynamic convolution \cite{chen2020dynamic} are less effective in lower model layers. Depth-based convolutions have been primarily designed for tasks like semantic segmentation \cite{wang2018depth, chen2021spatial, cao2021shapeconv} or 3D monocular object detection \cite{ding2020learning}. MobileNetV2 \cite{ni2024improved} reduces depth image noise and enhances features, but still needs improvement in challenging environments with small objects, low light, and clutter.  Involution \cite{li2021involution} reverses standard convolution to reduce inter-channel redundancy and enhance long-distance visual interactions with fewer parameters. Our research modifies involution to dynamically handle raw depth input. 
Table \ref{tab:conv_alternatives} summarizes these standard convolution alternatives. 

\begin{table}[t]
\centering
\caption{Summary of standard convolution alternatives and their main idea, and limitations.}
\label{tab:conv_alternatives}
\begin{tabular}{p{3.9cm} p{4.4cm} p{3.8cm}}
\toprule

\textbf{Method} & \textbf{Main Idea} & \textbf{Key Limitation} \\
\midrule

Standard Convolution \mbox{\cite{lecun1998gradient}} 
& Fixed spatial kernels 
& Limited adaptivity \\

Deformable Conv / DCNv2 \mbox{\cite{dai2017deformable,zhu2019deformable}} 
& Learnable geometric offsets 
& High computational cost \\

PAC \mbox{\cite{su2019pixel}} 
& Pixel-content adaptive filtering 
& Mainly effective for segmentation \\

CondConv / Dynamic Conv \mbox{\cite{yang2019condconv,chen2020dynamic}} 
& Input-dependent kernels 
& Weak in early layers \\

Depth-based Conv (Depth-aware, S-Conv, ShapeConv) \mbox{\cite{wang2018depth,chen2021spatial,cao2021shapeconv}} 
& Depth-guided spatial weighting 
& Not designed for detection \\

Transformer-based RGB-D \mbox{\cite{pan2024multi}} 
& Cross-modal attention 
& Memory and computation intensive \\

Depth Interaction Transformers \mbox{\cite{depth2024dipformer}} 
& Explicit depth–RGB interaction 
& Segmentation-oriented \\

Involution \mbox{\cite{li2021involution}} 
& Spatial-specific, channel-agnostic 
& Depth-agnostic \\

\bottomrule 
\end{tabular}
\end{table}

\begin{figure*}[t]
	\center
	\includegraphics[width=1\linewidth]{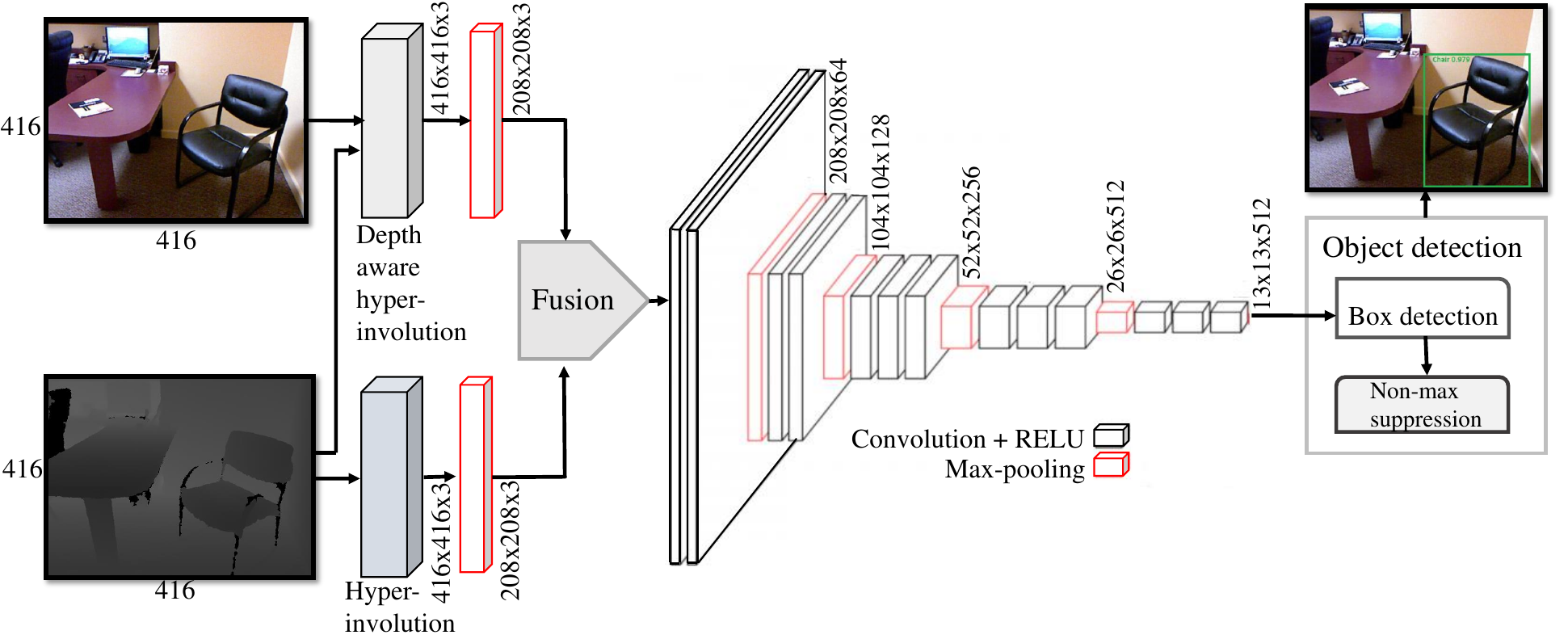}
	\caption{The proposed two streams and single stage detection architecture for real-time applications.}
	\label{MainModel}
\end{figure*}


\section{The Model} 
In this section, we first introduce the main RGB-D detection
architecture. Then we discuss the two main modules namely
the depth aware hyper-involution and fusion designed specifically for RGB-D detection.
\subsection{Two Streams Architecture}
Existing RGB-D object detection models typically use two-stage architectures, which hinder real-time performance. Our solution introduces a single-stage detector that predicts bounding boxes in a single pass, eliminating the need for a separate sparse prediction stage. As shown in Figure \ref{MainModel}, the model takes a color image and its corresponding depth map as input to two different network streams. One stream, containing the depth-aware hyper-involution (described in Section \ref{sec:depthhyper}) followed by a pooling layer, extracts color image features while paying parallel attention to depth. The second stream processes complementary semantic features from the depth map using a hyper-involution (described in Section \ref{hypersec}),
followed by a pooling layer to align shapes before fusion.
The features from the two streams are then combined using the fusion stage described in Section \ref{sec:fusionstage}. The post-fusion model consists of a backbone network with 13 convolutional layers, as shown in Figure \ref{MainModel}, inspired by \cite{simonyan2014very}. This backbone uses 3x3 convolutional layers with a stride of 1, consistent padding, and 2x2 max-pooling layers with a stride of 2, significantly reducing computational complexity. The final detection stage includes a detection head that provides classification and localization predictions through a non-max suppression layer. We use the loss function suggested by \cite{redmon2017yolo9000} due to its compatibility with this model's output and its success in state-of-the-art single-stage detector \cite{huang2018yolo}.

\subsection{Depth Aware Hyper-involution}\label{sec:depthhyper}
\begin{figure}
	\center
	\includegraphics[width=0.7\linewidth]{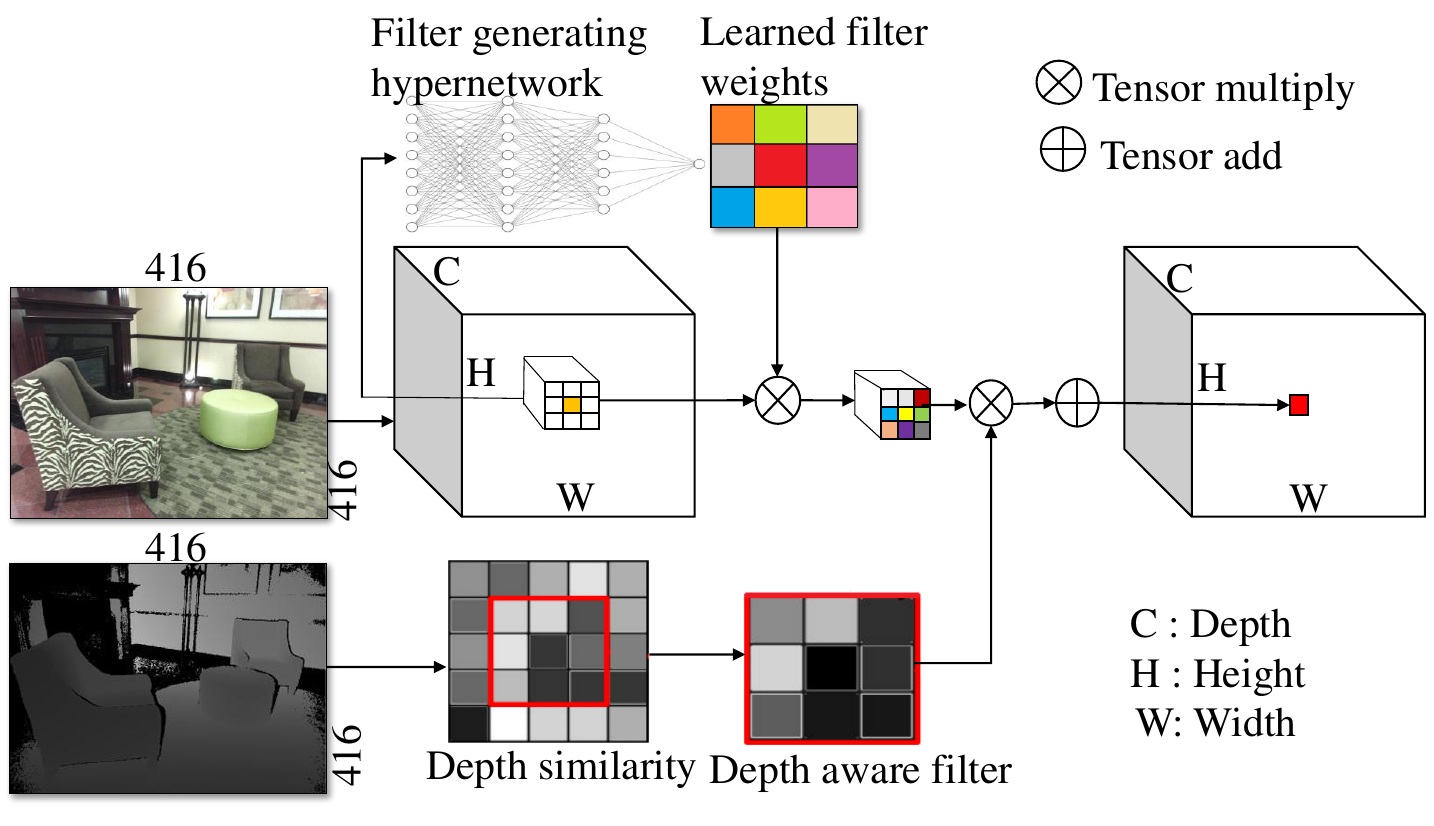}
	\caption{Depth Aware Hyper-involution: Depth similarity creates a depth-aware filter, with a hyper-network generating weights for each image region.}
	\label{Depth Aware Hyper-involution}
\end{figure}

The depth-aware hyper-involution module (Figure \ref{Depth Aware Hyper-involution}) is designed to incorporate spatial and depth information when processing color image features. To grasp its concept, it's essential to differentiate between convolution and involution.  

\textbf{Involution} The concept of involution~\cite{li2021involution} offers an alternative to standard convolution~\cite{lecun1998gradient} by incorporating spatial-specific and channel-agnostic features. It employs a generalized self-attention mechanism, enabling it to focus on specific image regions and capture long-range dependencies. This enhances its capability to model complex spatial relationships, making it potentially more effective for image-processing tasks. Additionally, its channel-agnostic nature efficiently reduces parameters while preserving the ability to capture intricate visual patterns. Precisely, an involution kernel of size $F\times F$ can be denoted as $\mathcal{H} \in \mathbb{R}^{H\times W\times F\times F\times G}$ where $G$ indicates the group of channels ($C$) in the input tensor that shares the same involution kernel. When such involution kernels undergo element-wise multiplication and addition on the image tensor, the final output feature tensor can be defined as in Equation \ref{eq:1.01}.
\begin{equation}\label{eq:1.01}
	O_{i,j,k} = \sum_{m= \lfloor \frac{-F}{2} \rfloor}^{\lfloor \frac{F}{2} \rfloor}  \sum_{n= \lfloor \frac{-F}{2} \rfloor}^{\lfloor \frac{F}{2} \rfloor} \mathcal{H}_{m+\lfloor \frac{F}{2} \rfloor, n+\lfloor \frac{F}{2} \rfloor,  \lceil\frac{kG}{C} \rceil}^{i, j}I_{i+m, j+n, k}
\end{equation}
In Equation \ref{eq:1.01}, $\mathcal{H}^{i, j}$ represents the involution kernel which is dynamically sampled from pixel position $I_{i,j}$ in the input tensor. Therefore, unlike the fixed filter of convolution operation, the involution filter is dynamically generated based on each spatial position of the input images. 
This feature allows the involution operation to focus on each spatial position in the image individually. Plus, it uses the same filter for multiple channels, which means it needs fewer parameters than standard convolution and reduces memory use.

\begin{figure}
\centering
	\includegraphics[width=0.85\linewidth]{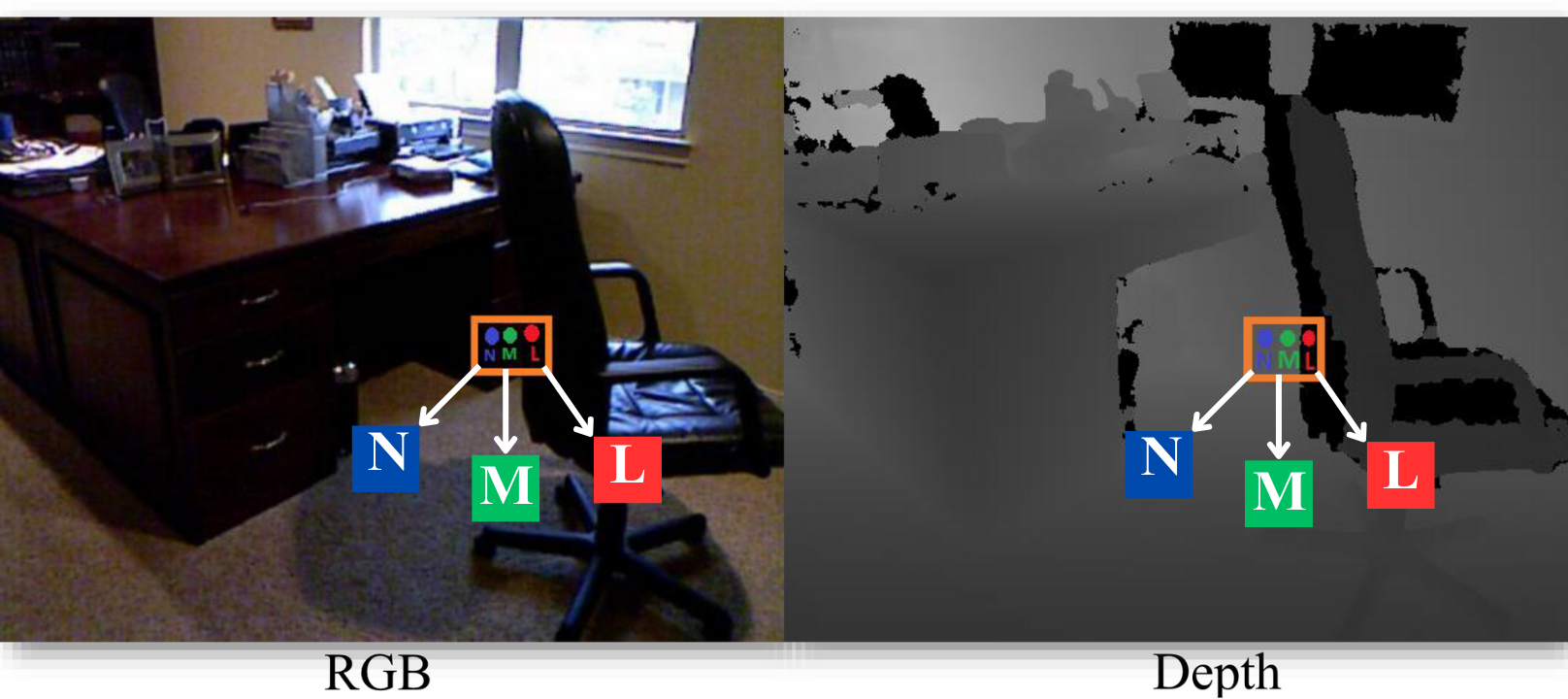}
	\caption[RGB and corresponding depth pixels]{Difference between pixels of RGB image and its corresponding depth map.}
	\label{Depth Similarity}
\end{figure}

\textbf{Depth Aware Involution} Nevertheless, involution was designed specifically considering the feature extraction from color images. It remains unaware of the depth of each pixel or spatial information while extracting features from the color image. For example, the RGB image 
highlights three pixels where pixels L, M, and N in Figure \ref{Depth Similarity} have the same pixel color as the chair and table have the same dark color. However, upon examining the depth map shown in Figure \ref{Depth Similarity}, it becomes clear that the depth of pixel L differs from that of pixels M and N. This is because the depth of pixel L is influenced by the chair, which is closer to the sensor than the part of the desk that pixels M and N correspond to.
To alleviate the effects of such unaccounted depth disparities in the detection accuracy, we redesign the involution operation to consider the spatial and geometric patterns from the depth map.
Given an input image tensor $I$ and depth map $D$, the output of our depth-aware hyper-involution operation is formulated as follows (Equation \ref{eq:2}).
\begin{multline}\label{eq:2}
	O_{i,j,k} = \sum_{m= \lfloor \frac{-F}{2} \rfloor}^{\lfloor \frac{F}{2} \rfloor}  \sum_{n= \lfloor \frac{-F}{2} \rfloor}^{\lfloor \frac{F}{2} \rfloor} \mathcal{P}_{m+\lfloor \frac{F}{2} \rfloor, n+\lfloor \frac{F}{2} \rfloor, \lceil \frac{kG}{C}\rceil}^{i, j} \mathbf{W}_{m+\lfloor \frac{F}{2} \rfloor, n+\lfloor \frac{F}{2} \rfloor}^{i, j} I_{i+m, j+n, k}
\end{multline}
In Equation \ref{eq:2} $\mathcal{P}^{i, j}$ represents the kernel that is dynamically generated via a new parameter-efficient filter generation hyper-network (described in a subsequent paragraph) which is conditioned on the pixel $I_{i,j}$. We select inverse multiquadric radial basis function as our depth weighing function. An RBF calculates a real number output solely based on the distance between the input and a constant reference point. The reference point can be either the origin or a specific center point \cite{buhmann_2003}. The specific RBF kernel is selected based on a quantitative comparison of different RBF kernels on benchmark and synthetic datasets as shown in Table \ref{tab:Weighing_options_new}. The inverse multiquadric kernel $\mathbf{W}_{m+\lfloor \frac{F}{2} \rfloor, n+\lfloor \frac{F}{2} \rfloor}^{i, j}$ captures the depth similarity between two pixels $D_{i,j}$ and $D_{i+m, j+n}$ and is defined as follows.
\begin{equation} \label{eq:3}
	\mathbf{W}_{p,q}^{i, j} = \frac{1}{\sqrt{1 + (\gamma \cdot (d(D_{i,j})-d(D_{p, q})))^2}}
\end{equation}
\begin{table}[h]
	\centering
         \caption[Comparison of Depth Weighting Functions.]{Accuracy comparison using different depth similarity weighting functions on three datasets.}
  \label{tab:Weighing_options_new}
        \begin{tabular}{p{3.1cm}p{2.0cm}p{2.0cm}p{1.0cm}}
			\hline
			
            \multicolumn{3}{c}{\textbf{Datasets}}& \\

			\textbf{Weighing Functions}  & NYU-Depth V2 & Sun RGB-D & Synthetic\\ 
   
			\hline
			  Triangular  &55.1 & 52.1&58.6 \\
			  Gaussian&53.3   &  52.4 & 58.7\\
			  Wendland&  55.1& 52.1 & 58.6\\
			 Inverse Multiquardic &55.4   & 52.7&58.9  \\
			\hline
			
	\end{tabular}
\end{table}

In Equation \ref{eq:3}, $d (D_{i, j})$ and $d(D_{p, q})$ denote the corresponding depth values at position $D_{i, j}$ and $D_{i+m, j+n}$, respectively. The choice of Equation \ref{eq:3} is based on the idea that the depth differences of various spatial locations and objects in the real scene should be addressed by using depth pixels from the depth map instead of solely relying on color that can often mislead like the one in Figure \ref{Depth Similarity}. The function decay rate is controlled by the parameter $\gamma$. The value of $\gamma$ is a constant that can be tuned until the detection model reaches the desired accuracy. In our case, the optimal value of $\gamma$ was 9.5 after testing in the range of 0.5 to 10 with an interval of 0.5. The decay parameter $\gamma$ controls the sensitivity of the inverse multiquadric RBF to depth differences. Based on RBF theory \mbox{\cite{buhmann_2003}}, meaningful $\gamma$ values typically lie within a moderate range; too small a value produces near-uniform weights, while too large a value makes the kernel oversensitive to depth noise. Therefore, we restricted $\gamma$ to the interval 0.5-10, which captures the theoretically stable operating region for this kernel. Within this range, we applied a coarse-to-fine grid search and observed consistent performance trends across datasets, with $\gamma$ = 9.5 yielding the best validation accuracy. This approach ensures that the chosen value is not a local optimum but the result of systematic evaluation across the full meaningful parameter space. More importantly, Equation \ref{eq:3} calculation does not add any extra parameter to Equation \ref{eq:2}. Furthermore, it is important to know that almost all the RGB-D datasets used for this research rely on different existing algorithms to deal with missing depth pixel values. For example, NYU Depth v2 uses an in-painting algorithm \cite{levin2004colorization} while SUN RGBD uses a different depth map improvement algorithm to estimate missing depth values \cite {song2015sun}. Therefore, this equation is not affected by missing depth pixels. Note that the hyper-involution shown in Figure \ref{MainModel} in our main object detection algorithm uses the same filter generation technique as the depth aware hyper-involution. However, it omits the depth aware component $\mathbf{W}_{m+\lfloor \frac{F}{2} \rfloor, n+\lfloor \frac{F}{2} \rfloor}^{i, j}$ since it is specifically used to extract complementary semantic features from the depth map. 

\begin{figure}[t]
	\center
	\includegraphics[width=0.9\linewidth]{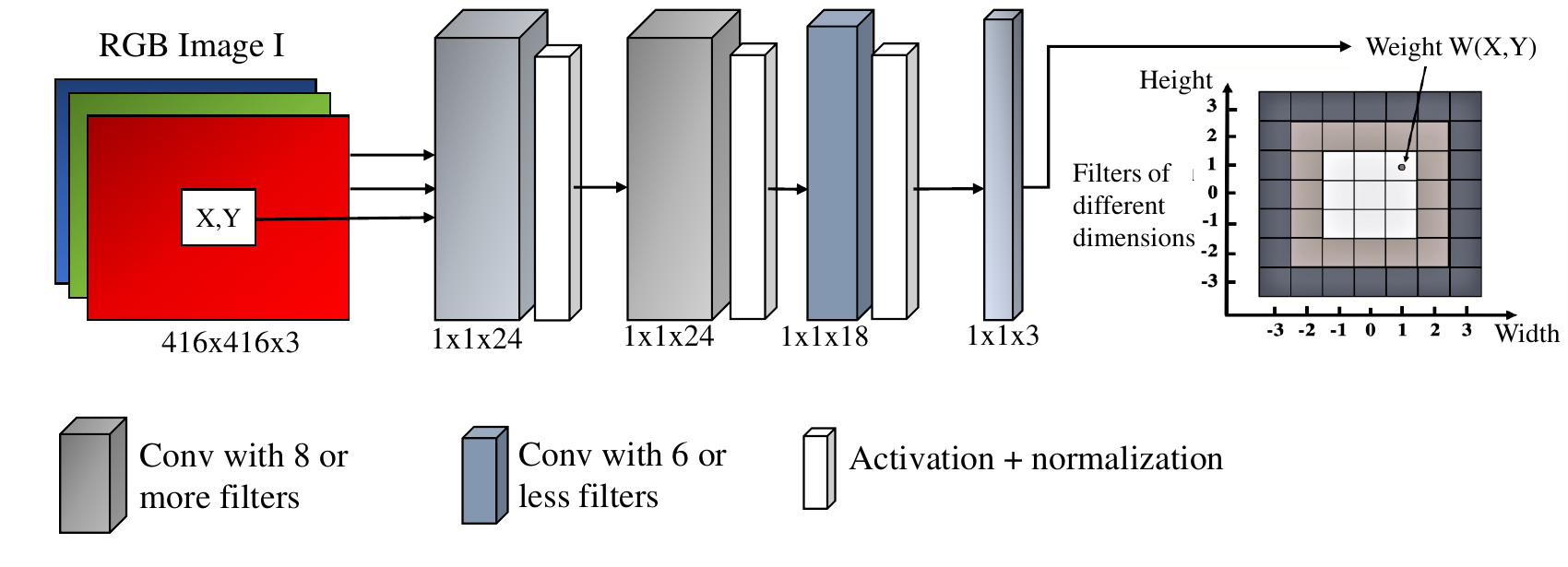}
\caption[Filter Generation Hyper-network]{The filter generation hyper-network learns filter weights for each RGB pixel individually}
	\label{Hypernetwork}
\end{figure}

\textbf{Filter Generation Hyper-network}\label{hypersec} We utilize a new function to map each 2D input kernel coordinate to the kernel value as shown in Figure \ref{Hypernetwork}. The function is a parameter efficient hyper-network.  The depth aware hyper-involution kernel weights are thus generated by a neural network (hyper-network) instead of independent learning. The trained weights of the kernel of a specific spatial location $\theta_{ij}$ can be represented as in \ref{eq:4}.
\begin{equation} \label{eq:4}
	\theta_{i,j} = N_2 \cdot \lambda(N_1 \cdot X_{i,j})
\end{equation}
In Equation \ref{eq:4}, $N_1$ and $N_2$ represent two linear transformations that collectively constitute a hyper-network. $N_1$ is implemented via 3 layers of 1$\times$1 convolution where the first two layers contain 8 filters with non-linear activation functions and the last layer consists of 6 filters. Meanwhile, $N_2$ is implemented using a single filter of 1$\times$1 convolution followed by a broadcasting of the output based on the size of the kernel. \(N_1\) processes the input element \(X_{i,j}\) through a series of convolutions and non-linear activations, while \(N_2\) applies a simpler convolution and broadcasting operation. $\lambda$ implies batch normalization and non-linear activation functions that interleave two linear projections. The main advantage of using this hyper-network in our depth aware hyper-involution is that the number of trainable parameters remains independent of the choice of the kernel size which is not possible in involution \cite{li2021involution} and standard convolution \cite{lecun1998gradient}. Thus, the expressiveness of our depth aware hyper-involution can be increased with a larger kernel size while keeping the number of trainable parameters constant. Note that the hyper-network used in \cite{ma2022hyper} is also independent of kernel size but it still depends on the number of input channels, output channels, and number of nodes in the final layer of their hyper-network. Whereas our hyper-network does not rely on the number of channels or number of nodes as these values remain constant.
\subsection{Fusion Stage} \label{sec:fusionstage}

\begin{figure}
        \centering
	\includegraphics[width=0.65\linewidth]{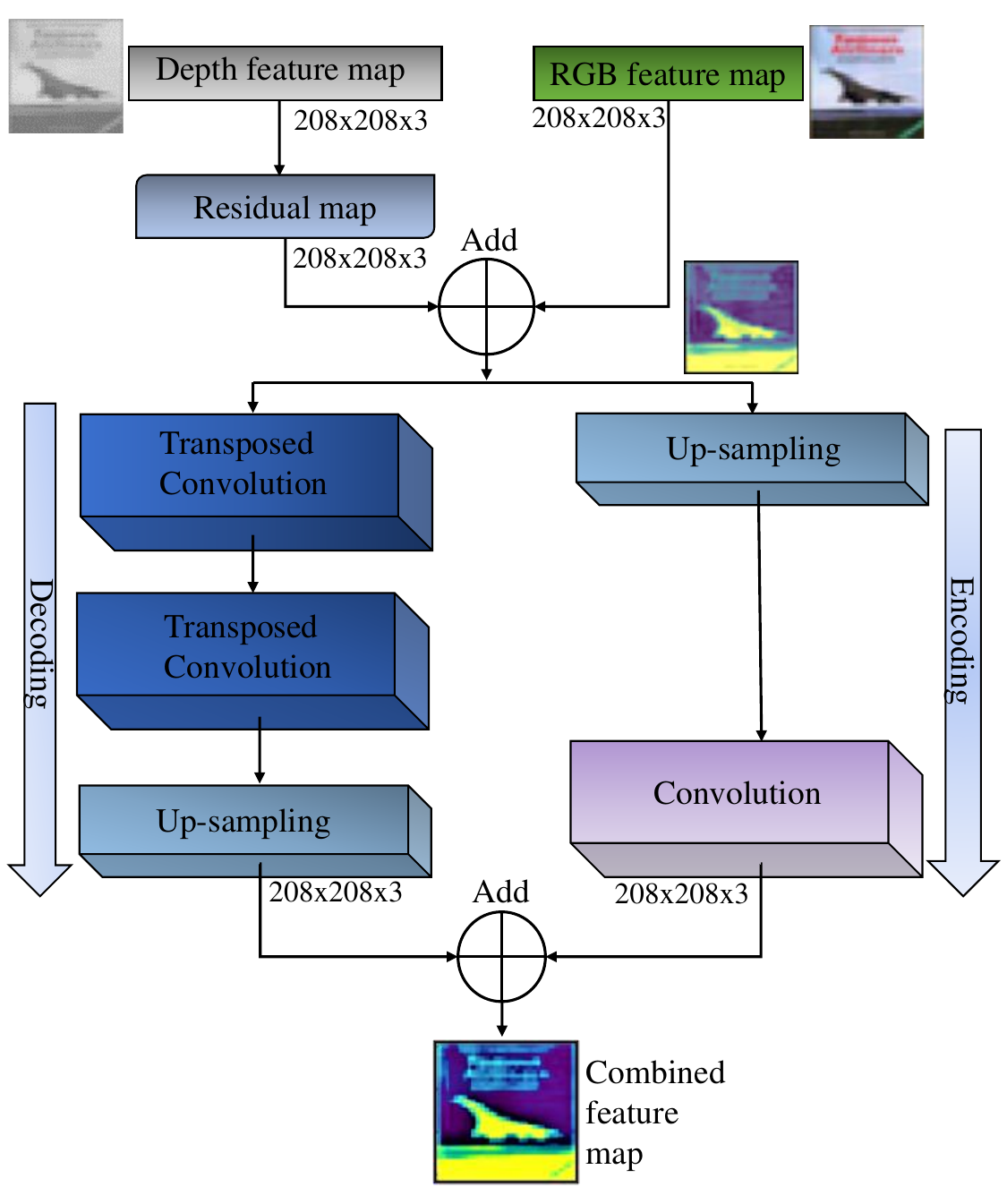}
	\caption[Fusion stage mechanism]{Illustration of the RGB-depth fusion stage, where depth-derived residual features are integrated with RGB features and refined through encoding-decoding operations to generate a combined feature map.}
	\label{Fusion}
\end{figure}

The fusion stage merges color and depth features from separate neural network streams, one focusing on semantic details from the depth map and the other on color image features. Preserving information from both streams is crucial. Previous methods often hindered feature flow and used basic concatenation without trainable parameters, limiting adaptability. Our approach enables parallel training to reduce information loss and effectively combine both streams. In our fusion module (Figure \ref{Fusion}), we use residual mapping to address modality-specific differences, transforming the depth feature map for element-wise addition with the RGB feature map. This merges depth and RGB feature tensors but lacks trainable weights for dynamic training. We enhance this with an encoder-decoder structure, inspired by semantic segmentation models
\cite{siddique2021u,zhou2018unet++}, enriching feature representation. The encoder processes the added feature tensor with convolutions, while the decoder enhances visual representation using transposed convolution to increase input dimensions efficiently. The final addition incorporates encoded information from the encoder into the decoder, preserving information and generating a fine-grained feature map. Trainable weights in the encoder and decoder aid fusion stage training alongside detector training.

\begin{figure*}
	\center
	\includegraphics[width=1\linewidth]{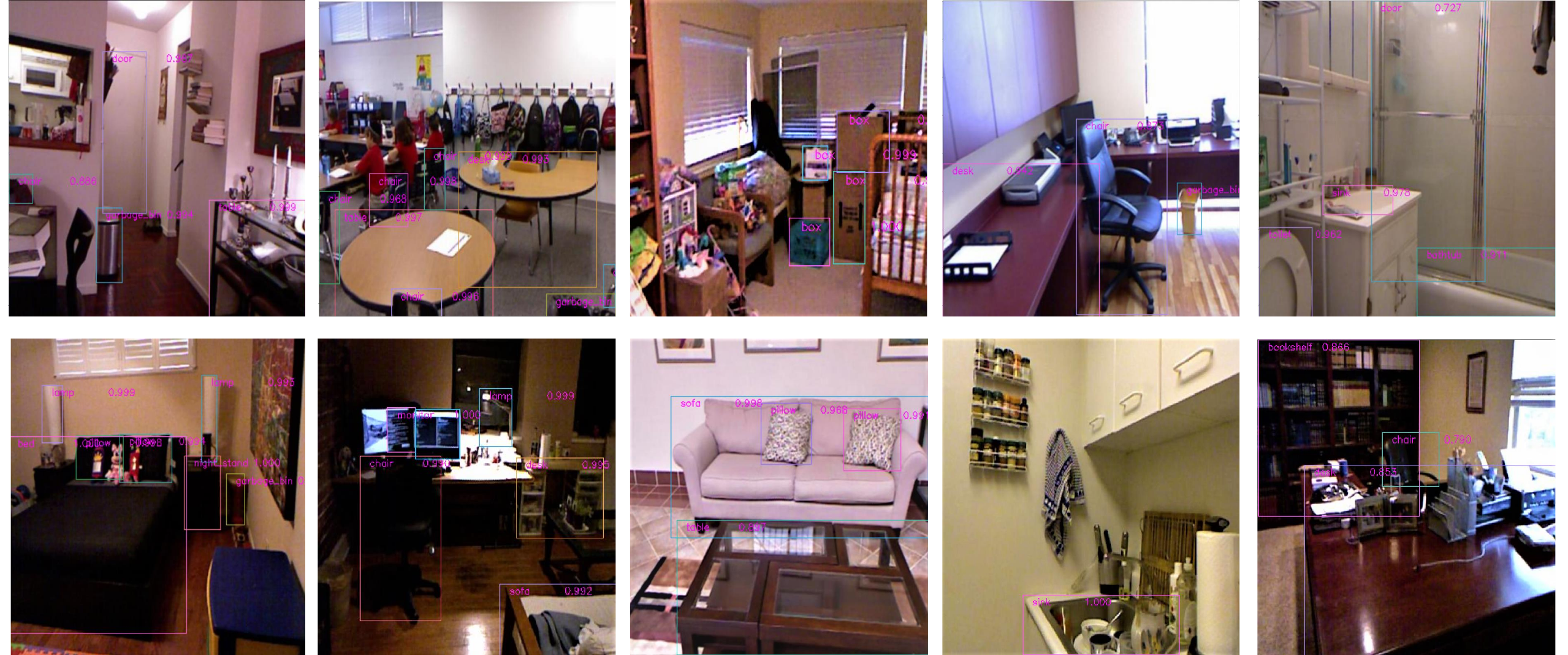}
	\caption{Top five images show detection results on SUN RGB-D; bottom images show detections on NYU Depth v2.}
	\label{Detection_result1}
\end{figure*}

\subsection{Implementation Details}
Our RGB-D detection model was implemented in Tensorflow version 2.5 and trained on a remote server provided by ACENET Canada with an NVIDIA Quadro RTX 6000 GPU and 24 GB memory. MATLAB was used to decode compressed datasets, and a Python script used dataset folders to fit our model's input requirements. As suggested by \cite{li2018cross},  
we select 19 furniture classes for object detection: bathtub, bed, bookshelf, box, chair, counter, desk, door, dresser, garbage bin, lamp, monitor, nightstand, pillow, sink, sofa, table, television, and toilet. 
For training the RGB-D object detection model, we used the Adam optimizer without pre-trained ImageNet weights. Input images were resized to 415 $\times$ 415 pixels. We trained the SUN RGB-D and NYU Depth V2 datasets for 150 and 130 epochs, respectively, with a learning rate of 0.0005. Similarly, the outdoor RGB-D dataset was trained for 160 epochs with the same learning rate. Synthetic data were trained for 120 epochs with a learning rate of 0.00009. Nonmax suppression was applied with an IOU threshold of 0.5. We utilized the loss function from \cite{redmon2017yolo9000}, considering localization, classification, and confidence losses for our single-stage detector.

\section{Experiments}\label{sec_exp}
In this section, we first present the performance of the model on benchmark datasets. We then present an overview of the proposed outdoor dataset and the model’s performance, followed by an ablation study analyzing the contributions of individual components. We evaluate our RGB-D object detection model using the benchmark NYU Depth v2 \cite{Silberman:ECCV12} and SUN RGB-D \cite{song2015sun} datasets. The official training test split guideline is followed for both of these datasets. To further explore the capacity of our model, we also use the synthetic RGB-D data generated by the automated pipeline (ref. Online resource) containing around 16000 RGB-D images. As customary with all other object detection research, mean average precision (mAP) and average precision (AP) are used as evaluation metrics, following the same technique proposed by PASCAL VOC \cite{Everingham15}.

\begin{table*}[h]
\centering
\caption{Experimental results on NYU Depth v2 \cite{Silberman:ECCV12} and SUN RGB-D \cite{song2015sun} with top three results highlighted in green, blue, and red, respectively. }
\label{tab:nyu_sunrgb_results}
\resizebox{1\linewidth}{!}{
\begin{tabular}{l l c c c cc}
\hline
\textbf{Method} & \textbf{Reference} & \textbf{Input modality} &
\textbf{RGB backbone} & \textbf{Depth backbone} &
\multicolumn{2}{c}{\textbf{mAP}} \\
\cmidrule(lr){6-7}
 &  &  &  &  &
\textbf{NYUv2} & \textbf{SUN-RGBD} \\
\hline
YOLOv8 \cite{yolov8}     & Ultralytics 2023 & RGB   & YOLOv8x & -- & 20.6 & 32.4 \\
YOLOv11 \cite{yolo11}     & Ultralytics 2024 & RGB   & YOLOv11x & -- & 21.4 & 32.9 \\
YOLOv12 \cite{yolo12}     & NeurIPS 2025 & RGB   & YOLOv12x & -- & 18.9 & 32.7 \\
Sparse R-CNN \cite{Sun_2021_CVPR}     & CVPR 2021 & RGB   & ResNet-50 & -- & 40.2 & 43.4 \\
ATSS \cite{zhang2020bridging}         & CVPR 2020 & RGB   & ResNet-50 & -- & 44.4 & 47.3 \\
GFL \cite{li2020generalized}          & NeurIPS 2020 & RGB & ResNet-50 & -- & 44.2 & 47.5 \\
SABL \cite{wang2020side}              & ECCV 2020 & RGB   & ResNet-50 & -- & 48.6 & 50.4 \\
Cascade R-CNN \cite{cai2018cascade}   & CVPR 2018 & RGB   & ResNet-50 & -- & 49.3 & 51.5 \\
Dynamic R-CNN \cite{zhang2020dynamic} & ECCV 2020 & RGB   & ResNet-50 & -- & 49.2 & 52.2 \\
Faster R-CNN \cite{ren2015faster}     & NeurIPS 2015 & RGB & ResNet-50 & -- & 49.7 & 52.3 \\
\hline
RGB-D R-CNN \cite{gupta2014learning}  & ECCV 2014 & RGB-D & AlexNet   & AlexNet   & 32.5 & 35.2 \\
Super Transfer \cite{gupta2016cross} & CVPR 2016 & RGB-D & VGG-16    & AlexNet   & 49.1 & 43.8 \\
AC-CNN \cite{li2016attentive}        & TMM 2016  & RGB-D & VGG-16    & AlexNet   & 50.2 & 45.4 \\
CMAC \cite{li2018cross}              & TIP 2018  & RGB-D & VGG-16    & AlexNet   & 52.3 & 47.5 \\
FetNet \cite{xiao2021fetnet}         & BMVC 2021 & RGB-D & ResNet-50 & ResNet-152 &
\textcolor{red}{\textbf{54.0}} & \textcolor{blue}{\textbf{54.5}} \\
MCTNet \cite{mctnet}            & MMM 2022 & RGB-D & ResNet-50 & ResNet-18  &
\textcolor{blue}{\textbf{54.8}} & \textcolor{green}{\textbf{55.7}} \\
\hline
Ours & -- & RGB-D & YOLO & Depth-aware Modulation &
\textcolor{green}{\textbf{55.4}} & \textcolor{red}{\textbf{53.3}} \\
\hline
\end{tabular}
}
\end{table*}

\subsection{Performance on Benchmark Datasets}
Table~\ref{tab:nyu_sunrgb_results} summarizes the overall detection performance on the NYU Depth v2 and SUN RGB-D benchmarks. While a growing number of recent RGB-D approaches adopt transformer-based architectures, many do not report results on these benchmarks under comparable evaluation protocols. To ensure a fair and reproducible comparison, we report results for widely adopted RGB-only and RGB-D baselines in Table~\ref{tab:nyu_sunrgb_results}, covering both classical and recent detectors. In addition to RGB-D methods, strong RGB-only single-stage and two-stage detectors are included to explicitly assess the contribution of depth information. Please note that te results in rows 4-16 of Table~\ref{tab:nyu_sunrgb_results} are taken from prior work and follow the training procedures reported in~\cite{mctnet}. The latest YOLO variants~\cite{yolov8,yolo11,yolo12} are trained using their recommended default configurations (100 epochs with a batch size of 16), consistent with common practice in recent literature.

Overall, the results demonstrate that incorporating depth information leads to consistent performance gains over RGB-only baselines. On NYU Depth v2, our method achieves the best performance with an mAP of 55.4\%, outperforming previously reported state-of-the-art approaches by a clear margin. On the more challenging SUN RGB-D benchmark, our approach attains the third-best overall performance, while remaining superior to all RGB-only detectors. This highlights the effectiveness of the proposed depth-aware feature modulation under heterogeneous indoor sensing conditions. The relatively smaller performance gap on SUN RGB-D can be attributed to architectural differences among RGB-D fusion strategies. Methods such as FetNet and MCTNet employ repeated or multi-level cross-modal fusion mechanisms, enabling stronger RGB--depth interactions across multiple network stages. In contrast, our approach adopts a single-stage fusion design to prioritize computational efficiency and architectural simplicity. While this design choice yields strong and competitive performance, it may limit the model’s ability to fully address the increased sensor variability and RGB-depth misalignment present in the SUN RGB-D dataset.

The class wise accuracies reported in Table \ref{tab:table_1.0} indicate that the proposed model significantly improves detection accuracy for classes like bed, monitor, desk, and toilet. However, lower accuracy with some objects may be due to occlusion and noisy depth maps, as our model relies heavily on depth information. Table \ref{tab:table_1.0} also reports the object detection accuracies of various models on SUN RGB-D.  FetNet \cite{xiao2021fetnet} scores higher over the proposed model on the counter class.The slightly lower performance of our method on SUN RGB-D can be explained by two main factors. First, FetNet employs three sequential feature-exchange modules within its RGB backbone, which reinforces cross-modal representations more aggressively than our single fusion module. Second, SUN RGB-D combines depth maps from four different RGB-D sensors, resulting in heterogeneous noise levels and occasional misalignment with the RGB image. Because our depth-aware hyper-involution directly incorporates depth similarity into kernel generation, such inconsistencies can negatively affect categories with irregular shapes or frequent occlusion (e.g., counter, box, lamp). These factors collectively account for the marginal performance difference on SUN RGB-D. A potential factor that may be working in favor of FetNet\cite{xiao2021fetnet} would be the unique feature exchange module which has been applied thrice in FetNet's RGB backbone network. The heterogeneity of objects within the box class, from small cereal boxes to large packages, challenges accurate detection, resulting in lower accuracy for this class. Furthermore, the desk class in the object detection benchmark \cite{gupta2014learning} is facing an issue with accuracy due to ambiguous data. Distinguishing between desks and tables is challenging, but our model is more accurate for desk class than others. Classifying lamps is tough due to light obscuring shapes in RGB images. Inconsistent depth data from four sensors in the SUN RGB-D dataset and mismatches between depth maps and RGB images can affect accuracy, as depth-aware hyper-involution relies on both types for learning filter weights.
Figure \ref{Detection_result1} visualizes our detection from these two datasets for qualitative evaluation. 
\begin{table*}
	\center
    \caption{Class-wise 2D object detection accuracy (AP, \%) on NYU Depth v2 and SUN RGB-D datasets. Results are reported per object category for RGB-D methods.}
\label{tab:table_1.0}
	\resizebox{1\linewidth}{!}
	{
		\begin{tabular}{p{1.4cm}p{0.8cm}p{0.8cm}p{0.8cm}p{0.8cm}p{0.8cm}p{0.8cm}|p{0.8cm}p{0.8cm}p{0.8cm}p{0.7cm}p{0.8cm}p{0.8cm}}
	\hline

        \multicolumn{1}{c}{} & \multicolumn{6}{c} {NYU Depth v2} & \multicolumn{6}{c} {SUN RGB-D}
        \\
        \hline
        
			Classes & \scriptsize {\textbf{RGB-D RCNN}} \cite{gupta2014learning} & \scriptsize {\textbf{Super Transfer}} \cite{gupta2016cross} & \scriptsize {\textbf{AC-CNN}} \cite{li2016attentive} & \scriptsize {\textbf{CMAC}} \cite{li2018cross} & \scriptsize {\textbf{FetNet}} \cite{xiao2021fetnet} & \scriptsize {\textbf{Ours}} 

   & \scriptsize {\textbf{RGB-D RCNN}} \cite{gupta2014learning} & \scriptsize {\textbf{Super Transfer}} \cite{gupta2016cross} & \scriptsize {\textbf{AC-CNN}} \cite{li2016attentive} & \scriptsize {\textbf{CMAC}} \cite{li2018cross} & \scriptsize {\textbf{FetNet}} \cite{xiao2021fetnet} & \scriptsize {\textbf{Ours}} \\
   
            \hline
            bathtub & 22.90 & 50.60 & 52.20 & 55.60 & 56.40  & 53.30 & 49.60 & 65.30 & 65.80 & 69.00 & 62.50  & 63.98 \\
            \hline
            lamp & 29.30 & 42.50 & 42.90 & 45.00 & 50.80  & 49.50 & 22.00 & 32.10 & 33.80 & 35.60 & 65.00  & 61.29 \\
            \hline
            bed & 66.40 & 81.00 & 82.40 & 83.90 & 78.30  & 94.09 & 76.00 & 83.00 & 83.30 & 86.10 & 80.90  & 81.42 \\
            \hline
            monitor & 43.60 & 62.90 & 63.60 & 65.80 & 69.50  & 73.37 & 10.80 & 36.80 & 39.50 & 40.50 & 43.10  & 50.46 \\
            \hline
            bookshelf & 21.80 & 52.60 & 52.50 & 54.00 & 57.30  & 52.40 & 35.00 & 54.40 & 56.20 & 57.90 & 47.90  & 53.45 \\
            \hline
            nightstand & 39.50 & 54.70 & 55.20 & 57.60 & 59.00  & 59.60 & 37.20 & 46.60 & 47.10 & 49.80 & 62.00  & 60.93 \\
            \hline
            box & 3.00 & 5.40 & 8.60 & 9.80 & 8.00  & 17.50 & 5.80 & 14.40 & 16.40 & 18.20 & 13.30  & 18.17 \\
            \hline
            pillow & 37.40 & 49.10 & 49.70 & 52.70 & 60.80  & 56.45 & 16.50 & 23.40 & 25.20 & 26.70 & 63.90  & 52.09 \\
            \hline
            chair & 40.80 & 53.00 & 54.80 & 55.40 & 68.20  & 69.46 & 41.20 & 46.90 & 47.50 & 50.30 & 69.30  & 63.10 \\
            \hline
            sink & 24.20 & 50.00 & 51.40 & 53.80 & 60.30  & 52.40 & 41.90 & 43.90 & 45.30 & 46.60 & 65.40  & 66.98 \\
            \hline
            counter & 37.60 & 56.10 & 57.30 & 59.20 & 37.60  & 54.34 & 8.10 & 14.60 & 16.00 & 17.40 & 49.20  & 17.80 \\
            \hline
            sofa & 42.80 & 65.90 & 66.80 & 69.10 & 69.00  & 69.50 & 42.20 & 61.30 & 61.90 & 67.20 & 56.30  & 57.90 \\
            \hline
            desk & 10.20 & 21.00 & 22.70 & 24.10 & 32.50  & 38.73 & 16.60 & 23.90 & 24.90 & 26.80 & 30.40  & 35.40 \\
            \hline
            table & 24.30 & 31.90 & 33.50 & 35.00 & 36.00  & 36.90 & 43.00 & 48.70 & 49.00 & 52.90 & 49.50  & 49.71 \\
            \hline
            door & 20.50 & 34.60 & 34.10 & 36.30 & 44.20  & 41.20 & 4.20 & 15.30 & 16.60 & 17.30 & 52.60  & 51.80 \\
            \hline
            tv & 37.20 & 50.10 & 51.80 & 56.90 & 55.40  & 55.46 & 32.90 & 50.50 & 54.10 & 56.70 & 40.30  & 39.18 \\
            \hline
            dresser & 26.20 & 57.90 & 58.10 & 58.50 & 59.10  & 53.70 & 31.40 & 41.30 & 42.70 & 44.40 & 41.90  & 40.23 \\
            \hline
            toilet & 53.00 & 68.00 & 70.40 & 74.70 & 71.20  & 72.50 & 69.80 & 79.40 & 84.20 & 84.90 & 85.50  & 83.42 \\
            \hline
            garbagebin & 37.60 & 46.20 & 46.50 & 47.20 & 51.90  & 52.20 & 46.80 & 51.00 & 53.40 & 54.40 & 56.90  & 54.00 \\
            \hline
            
            \hline
		\end{tabular}
	}
  
\end{table*}
\begin{figure*}
	\center
	\includegraphics[width=1\linewidth]{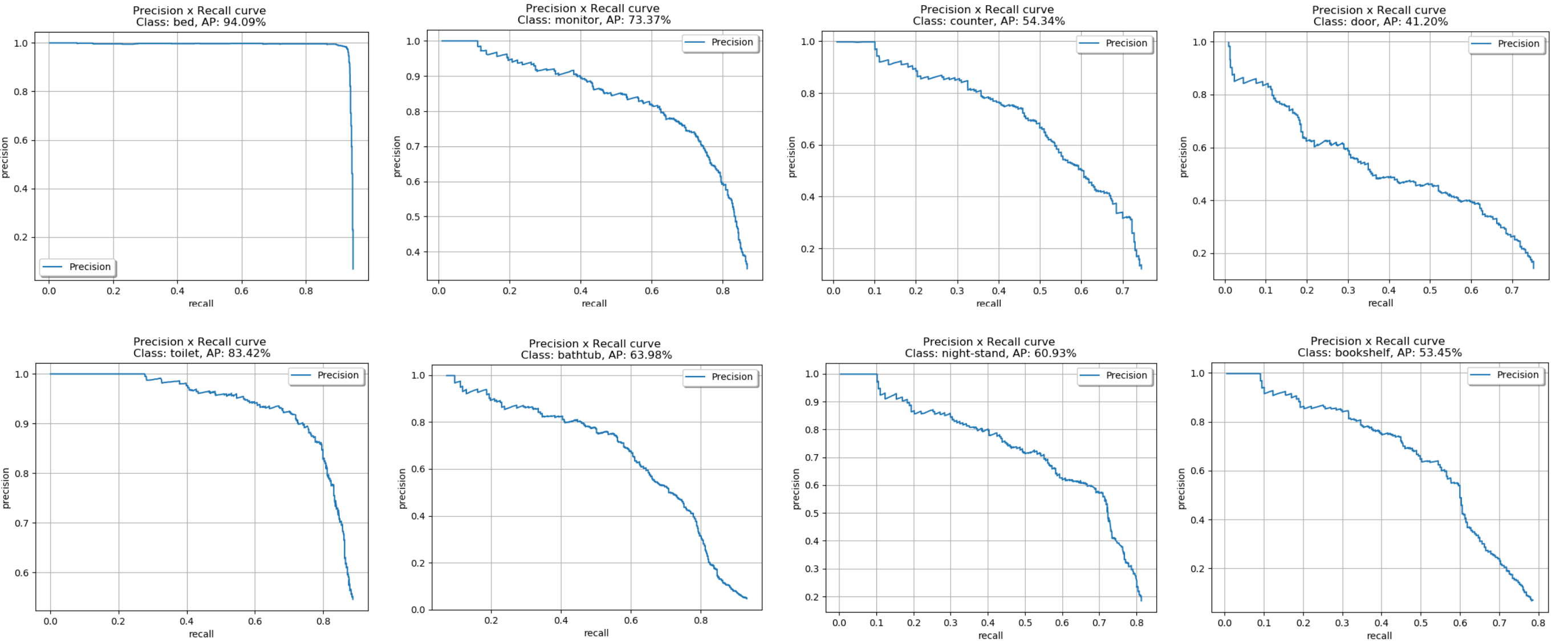}
	\caption[Precision-Recall Curves]{Precision recall curves for different classes on SUN RGB-D and NYU Depth v2. The top row shows classes in NYU Depth v2 while the bottom row shows classes in SUN RGB-D.}
	\label{PR_curves}
\end{figure*}

Figure \ref{PR_curves} shows class-wise precision–recall curves on NYU Depth v2 (top row) and SUN RGB-D (bottom row), offering insight beyond aggregate mAP scores into the behavior of the detector across object categories. On NYU Depth v2, classes such as bed and toilet maintain high precision over a wide recall range, reflecting their stable geometry and distinctive depth structure, which are effectively captured by the proposed depth-aware hyper-involution. In contrast, classes such as monitor, counter, and door exhibit a more gradual precision drop as recall increases, consistent with greater appearance variation, partial occlusions, and cluttered backgrounds. The SUN RGB-D curves generally show steeper precision degradation, aligning with the dataset’s heterogeneous sensor sources and varying RGB–depth alignment, which particularly affect categories such as bathtub, night-stand, and bookshelf that involve thin structures or planar ambiguities. Despite these challenges, the model maintains strong precision in the low-to-mid recall regime across both datasets, indicating that depth-guided feature modulation improves detection reliability under realistic indoor sensing conditions and complements the overall mAP improvements.

\begin{figure*}
	\center
	\includegraphics[width=0.9\linewidth]{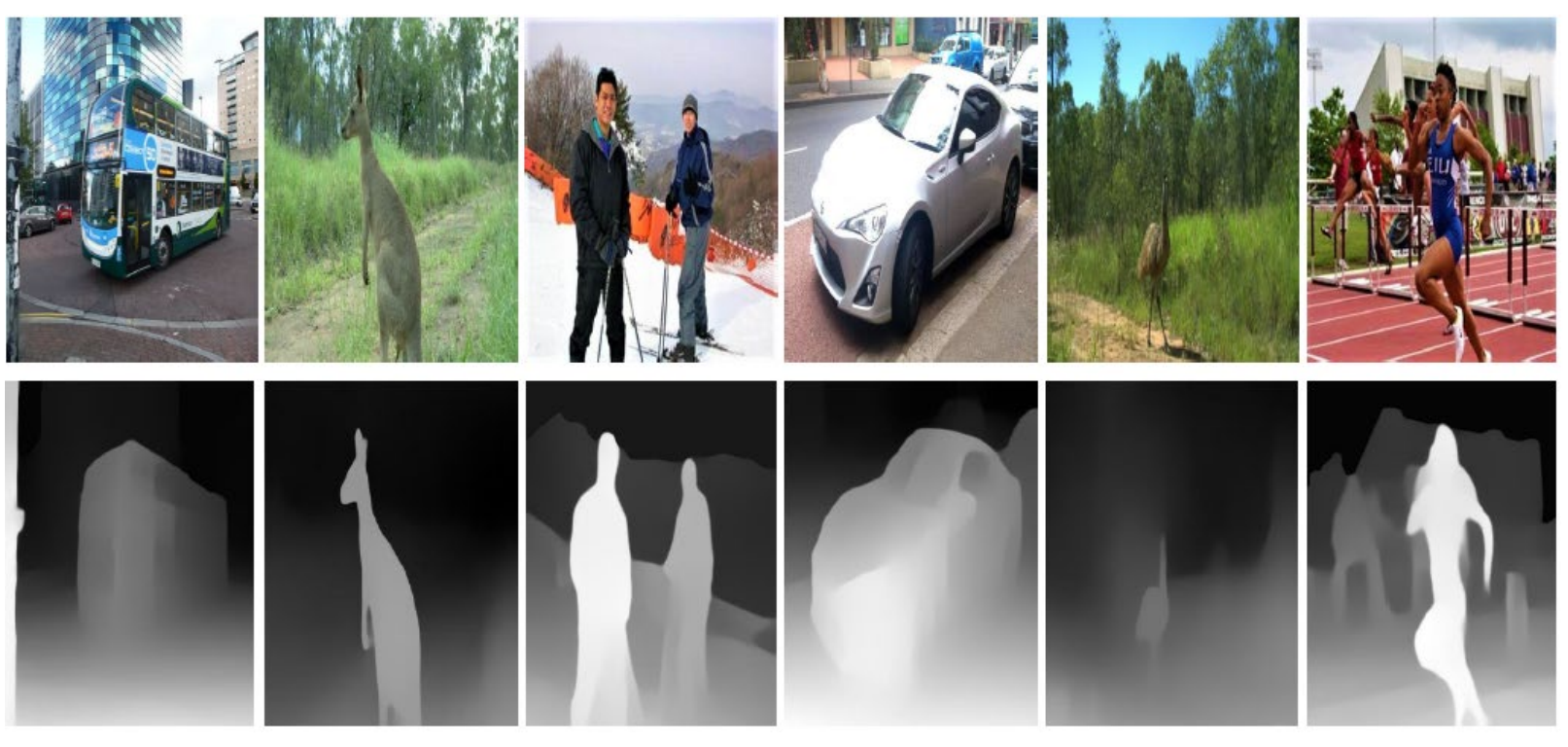}
	\caption[Outdoor RGB-D Dataset]{Some samples of RGB images and their corresponding depth map from the curated Outdoor RGB-D Dataset.}
	\label{Outdoor_RGB-D}
\end{figure*}

\subsection{Outdoor RGB-D Dataset} \label{sec_novel_data}
One of the significant limitations of the benchmark RBG-D object detection datasets used in the literature is that they contain RGB-D data only from indoor environments. This limitation leaves several questions unanswered for the research community, such as the performance of RGB-D object detection in challenging outdoor lighting conditions, or whether RGB-D data is only useful for indoor environments. To address this concern, we introduce a new fully annotated RGB-D dataset named the Outdoor RGB-D Detect dataset. This dataset exclusively consists of RGB-D image pairs captured in diverse outdoor environments. The RGB images in this dataset are sourced from three benchmark datasets: Places \mbox{\cite{zhou2017places}}, Open Images \mbox{\cite{OpenImages,OpenImages2}}, and the multi-class wildlife dataset \mbox{\cite{zhang2020omni}}. Depth maps for these images were predicted using the dense vision transformer (DPT-hybrid) \mbox{\cite{ranftl2021vision}}. The depth maps in the Outdoor RGB-D Detect dataset are generated using DPT-hybrid, which produces pseudo-depth rather than true sensor measurements. While such pseudo-depth may contain noise or scale ambiguity compared to sensor-based depth, our method relies on relative depth similarity instead of absolute depth values, reducing sensitivity to these errors. Moreover, existing benchmarks such as NYU Depth v2 and SUN RGB-D already employ different depth sensors and post-processing pipelines, indicating that robustness to heterogeneous depth sources is essential for RGB-D detection. Three object classes-Humans, Animals, and Vehicles are selected for detection, representing common outdoor elements. Despite the limited number of classes, the dataset is challenging due to the diverse subtypes within each class e.g., vehicle class has instances of bus, truck, SUV within Vehicle, while Animal class have images of Kangaroos, Ostrich, Dog, etc. Additionally, the outdoor environments exhibit wide variation, ranging from dense forests to busy downtown areas, with varying weather and lighting conditions. Some samples from this dataset are shown in Figure \ref{Outdoor_RGB-D}. The dataset comprises 1819 RGB-D samples, with 997 samples for training and 822 for testing, and importantly, it exhibits no class imbalance, unlike many existing benchmarks.
Bounding box annotations were created on RGB images using the MIT LabelMe framework \cite{russell2008labelme}. The annotated bounding boxes were then consistently transferred to the corresponding aligned depth maps to ensure precise spatial correspondence between modalities. The annotation process was carried out by a team of 13 trained annotators and subsequently reviewed and corrected by the authors to ensure annotation quality and reliable supervision for RGB-D object detection.

\subsection{Performance on Outdoor RGB-D Dataset}

\begin{figure}[h]
	\centering
	\includegraphics[width=1\linewidth]{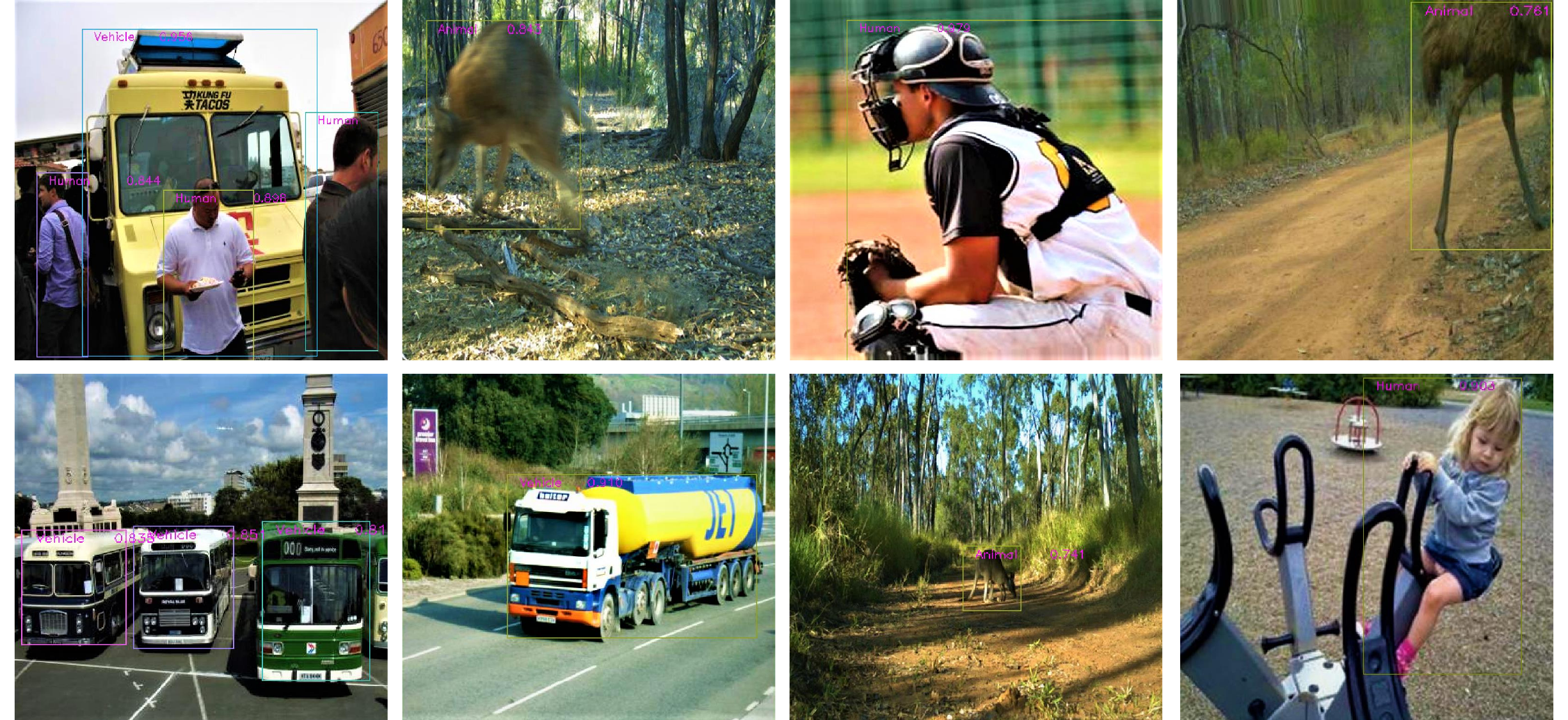}
	\caption[Detection results on Outdoor RGB-D dataset.]{A few detection results on the Outdoor RBG-D.}
	\label{Detection_result_outdoor}
\end{figure}

Figure \ref{Detection_result_outdoor} shows some of the qualitative detection results on our outdoor RGB-D detection dataset. This figure highlights the challenges of detecting objects in a dataset with just three classes, due to the variety within each class. For instance, the images in the first row and the first two images in the second row of Figure \ref{Detection_result_outdoor} show our detector identifying vans, buses, and trucks as vehicles despite their different appearances. This indicates that the detector effectively learned the diverse features within each class. It also successfully detected fast-moving objects, like the blurred one in row one, column two. The detection of a person wearing a helmet in the first row, third column, demonstrates the model’s ability to generalize. Additionally, spotting animals from a distance in a dense forest, as shown in row two, column three, further proves the model’s accuracy.

In the quantitative experiments as shown in Table \ref{tab:table_5.6}, our model achieved an mAP of 80.1 which is also significantly higher FETNet \cite{xiao2021fetnet}. Therefore, both qualitative and quantitative results indicate the high capacity of our detection model in real world outdoor environments under diverse lighting conditions.

\begin{table}	
	\centering
  \caption{Experimental results on the Outdoor RGB-D dataset.}
	 \label{tab:table_5.6}
	{
		\begin{tabular}
  {p{1.9cm}p{1.9cm}p{1.9cm}p{1.9cm}p{1.9cm}}
			\hline 
			Method & \textbf{mAP} & Vehicle & Human & Animal \\
			\hline
			
			FETNet \cite{xiao2021fetnet} & 78.4 & 79.5 & 77.7 & 78.1 \\
			\hline
			\textbf{Ours} & 80.2 & 81.1 & 80.7 & 78.8 \\
			\hline
		\end{tabular}
	}

\end{table}

\begin{figure}[h]
	\center
	\includegraphics[width=1\linewidth]{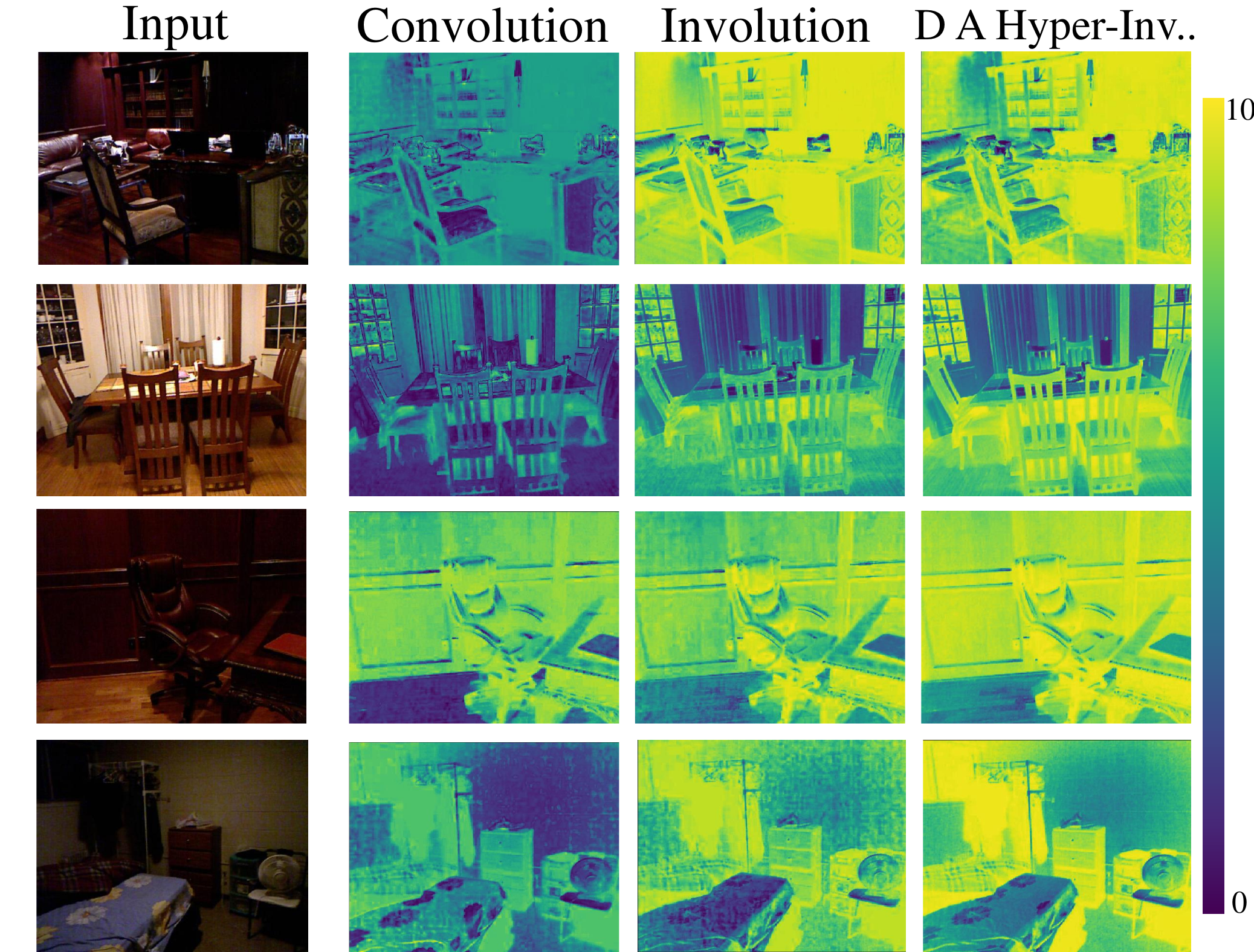}
 \caption{Heat maps for image instance from NYU Depth v2.}
	\label{Filter_heatmaps}
\end{figure}

\subsection{Feature Maps Analysis}

To visually analyze the trained depth-aware hyper-involution kernel, we sum the F $\times$ F values from each kernel (where F represents the kernel's height and width) as its representative value and compare it with similarly trained convolution and involution kernels. These representative values at various geometric positions form the heat maps shown in Figure \ref{Filter_heatmaps}. The columns following the input images in this figure represent the learned kernels of convolution, involution, and depth-aware hyper-involution, respectively.
From Figure \ref{Filter_heatmaps}, it is evident that depth-aware hyper-involution captures significant semantic features of input images by utilizing depth information. Specifically, the first row's last column shows that the depth-aware hyper-involution accurately maps the sharp edges of a bookshelf, even in dark areas, unlike convolution and involution. This suggests that depth-aware hyper-involution can highlight edges regardless of darkness using depth information. In the last column of the second row, depth-aware hyper-involution stands out by highlighting darker regions in yellow, which convolution and involution miss. It also maintains consistent color coding for objects at the same depth, like pairs of chairs. The last row’s fourth column maps flower textures on a bed more accurately than other filters. Additionally, in the last column of the third row, depth-aware hyper-involution preserves floor texture details better, showing its strength in spatial-specific features.

\begin{figure}
	\center
	\includegraphics[width=12.5cm]{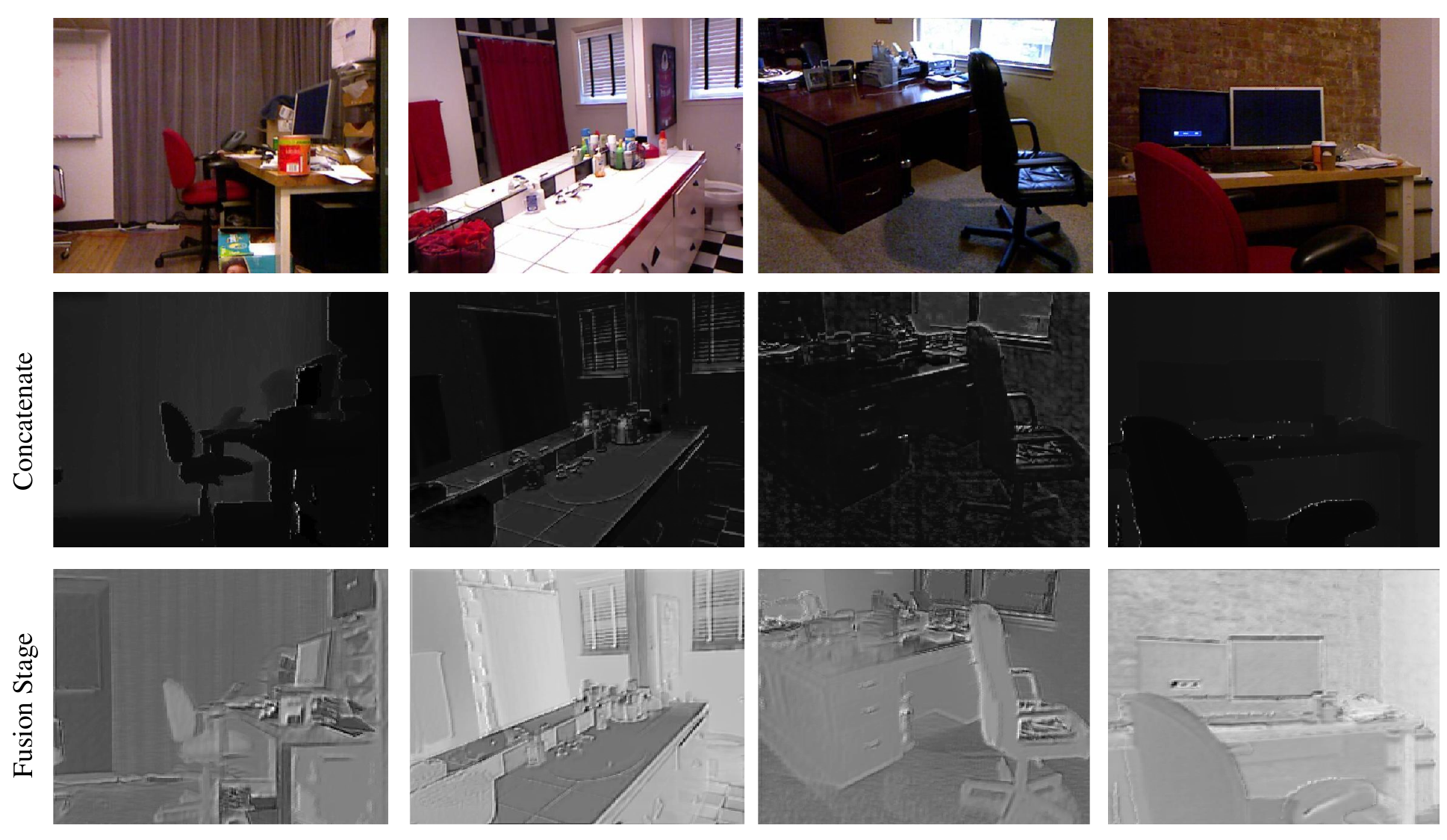}
	\caption[Visualization of fusion feature maps.]{Feature map outputs after concatenation and fusion stages for image samples from SUN RGB-D and NYU Depth v2.}
	\label{Fusion_featuremap}
\end{figure}

To compare fusion and concatenation, we visualize their output feature maps. Figure \ref{Fusion_featuremap} shows that fusion better combines and preserves details from both the image and its depth. For example, comparing the second and third rows of the first column in Figure \ref{Fusion_featuremap}, the wall with the whiteboard is visible in the fusion feature map but completely obscured in the concatenation output. Similarly, the second and third rows of the second column show that the fusion feature map captures the checkerboard texture of the wall behind the red curtains, a detail missed in the concatenation output. This visually supports the idea that using an encoder to encode rich semantic features and a decoder to upsample the combined feature map is effective. Another distinction can be seen by comparing the second and third rows of the third column, where the outer boundary of the chair and desk is visible in the fusion output but not in the concatenation output. Similarly, in the last column of the second and third rows, the fusion output preserves the boundary between two monitors, whereas the concatenation feature map merges them into one.

\subsection{Inference and Parameters}
\textbf{Inference Cost:} Inference GFLOPs refer to the number of floating-point operations needed for a single prediction on a trained model, measuring computational complexity and performance in GigaFLOPs ($10^9$ FLOPs). Calculating Inference GFLOPs involves counting the additions and multiplications required for each layer’s activations for a given input and converting that count to FLOPs. We compare the Inference GFLOPs of our detection model with several state-of-the-art RGB and RGB-D object detectors to assess computational performance. For RGB-D detectors, we use FETNet \cite{xiao2021fetnet} and other implementations reported in their paper. As shown in Table \ref{tab:table_4.0}, our detection model achieved the lowest Inference GFLOPs, indicating the least computational complexity. Our model also significantly outperforms the real-time single-stage detectors including the recent YOLO variants \cite{redmon2017yolo9000, yolov8, yolo11, yolo12} in GFLOPs, highlighting its real-time performance. A key reason for achieving lower Inference GFLOPs is our backbone structure, featuring 3x3 convolution layers with stride 1 and same padding, and maxpool layers with 2x2 filters and stride 2. The fusion layer mainly uses 3 filters for convolution operations. We use just 8 filters for depth-aware hyper-involution, resulting in impressive inference times.

\begin{table*}[h]
\centering
\caption{Inference GFLOPs comparison with state-of-the-art RGB and RGB-D based detection algorithms on SUN RGB-D.}
\label{tab:table_4.0}

\resizebox{0.7\textwidth}{!}{
\begin{tabular}{p{6cm} c c c}
\hline
\textbf{Model} & \textbf{RGB} & \textbf{RGB-D} & \textbf{GFLOPs} \\
\hline
YOLOv2 \cite{redmon2017yolo9000}               & \checkmark &              & 63.03 \\
YOLOv8x~\cite{yolov8}            & \checkmark &              & 258.5 \\
YOLOv11x~\cite{yolo11}            & \checkmark &              & 195.6 \\
YOLOv12x~\cite{yolo12}            & \checkmark &              & 200.3 \\
Cascade R-CNN \cite{cai2018cascade}           & \checkmark &              & 168.3 \\
Faster R-CNN \cite{ren2015faster}             & \checkmark &              & 140.5 \\
Cascade R-CNN + FEM + MViT \cite{xiao2021fetnet} &           & \checkmark   & 158.5 \\
Faster R-CNN + FEM + MViT \cite{xiao2021fetnet}  &           & \checkmark   & 130.7 \\
FETNet \cite{xiao2021fetnet}                  &           & \checkmark   & 279.3 \\
\hline
\textbf{Our model}                            &           & \checkmark   & \textbf{26.72} \\
\hline
\end{tabular}
}
\end{table*}

\begin{table*}[h]

\centering
\caption[Inference and mAP comparison plot.]{Inference, mAP and number of parameters comparison for model versions in ablation study.} 
\label{tab:inference_mAP_plot}    
\resizebox{0.65\columnwidth}{!} {

\begin{tabular}{llll}

\noalign{\smallskip}\hline\noalign{\smallskip} 
\textbf{Model Versions}                        & GFLOPs $\scriptstyle\downarrow$ & mAP $\scriptstyle\uparrow$ & \# parameters $\scriptstyle\downarrow$ \\ 
\noalign{\smallskip}\hline\noalign{\smallskip} 
Baseline         & 27.61        & 46.8   & 14816774          \\ 
Baseline+Fusion    & 26.99        & 49.7  &  14829214            \\ 
Inverse Multiquardic & 26.84        & 52.1 & 14809700             \\ \hline
Proposed Model & \textbf{26.72}        & \textbf{53.3}  & \textbf{14808688}            \\ \hline
\end{tabular}
}
   
\end{table*}

\subsection{Ablation Study}
\subsubsection{Module Test}
In the ablation study, the baseline was constructed by replacing the proposed depth-aware hyper-involution with standard convolution and substituting the fusion stage with simple feature-map concatenation within the same detection architecture. Subsequently, we replaced concatenation with fusion in the baseline to evaluate the proposed fusion's performance. As reported in Table \ref{tab:inference_mAP_plot}, the original model achieved the highest accuracy which verifies the usefulness of the depth aware hyper-involution. Our main detection model has the minimum inference GFLOPs when compared with the baseline and baseline with just fusion or depth aware-involution. This implies that the fusion stage and depth aware hyper-involution do not increase computational complexity and help to maintain the real-time performance of the detection model. Moreover, the model also has less parameters when compared to the model with only fusion and standard convolution which suggests the depth aware hyper-involution operation consumes less memory than standard convolution. When normal concatenation is replaced with the suggested fusion in the baseline model, the number of parameters increases significantly. This indicates that the fusion module has more trainable parameters, which can enhance the model's learning ability.

\subsubsection{Number of Parameters Vs Kernel Sizes}
\begin{figure}
	\includegraphics[width=0.95\linewidth]{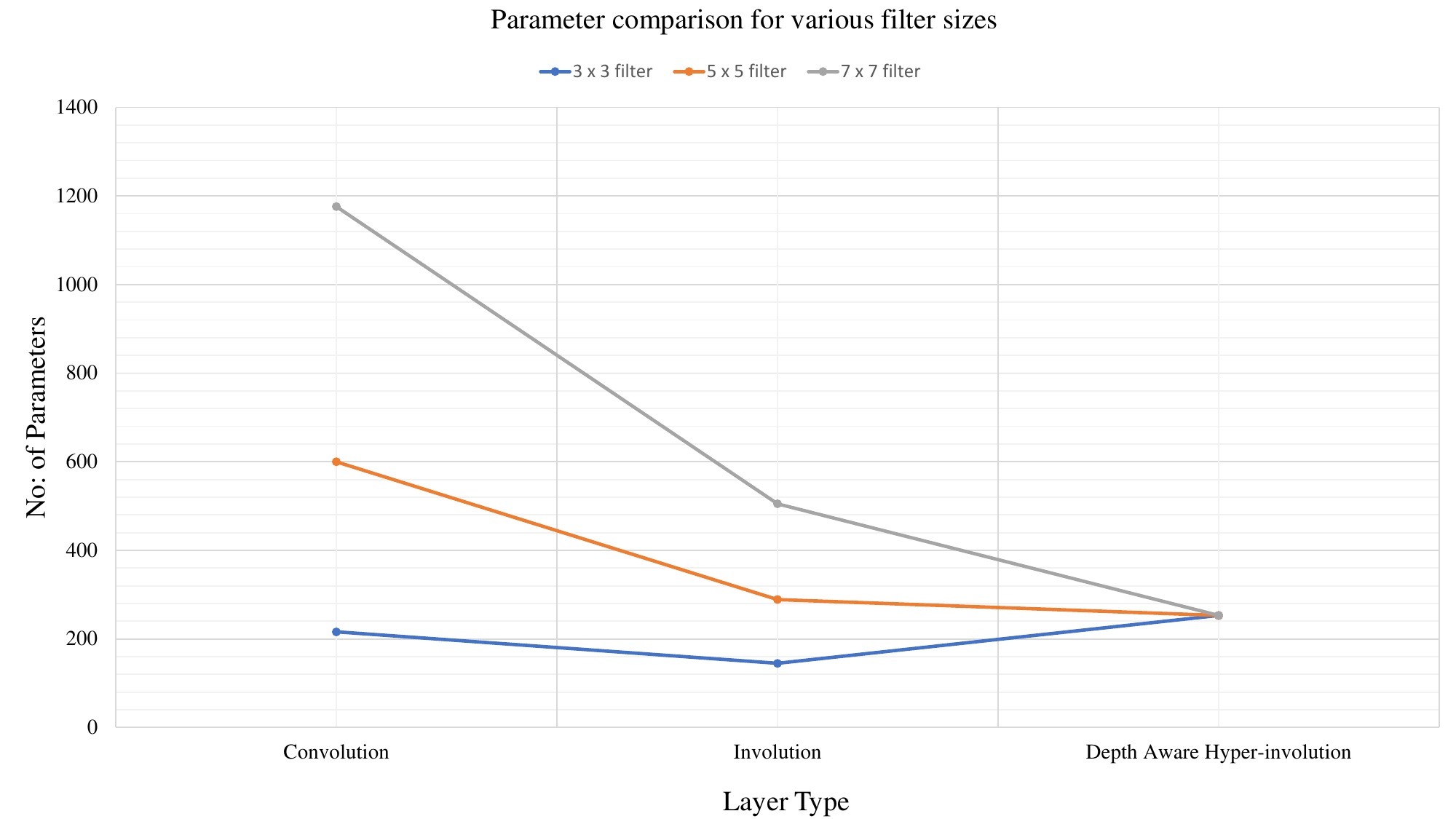}
	\caption[Kernel comparison plot.]{Parameter comparison for different kernel sizes of convolution, involution and our depth aware hyper-involution.}
	\label{Filter_comparison}
\end{figure}

Furthermore, we conducted another ablation study to see the effect on number of parameters of depth aware hyper-involution for different kernel sizes. We also compare it with the parameters of standard convolution \cite{lecun1998gradient} and involution \cite{li2021involution} for similar kernel sizes. As shown in Table \ref{tab:table_6.0} and the graph plot in \ref{Filter_comparison}, the parameters of depth aware hyper-involution remains the same for all kernel sizes which is not the case in involution and standard convolution. Moreover, the number of parameters in depth aware hyper-involution is less than that of standard convolution for all sizes of filters. This clearly indicates the usefulness of the hyper-network in generating filters for the depth aware hyper-involution. Note that, we applied 8 filters for all these modules during comparison.

\begin{table*}[h]
\centering
\caption{Parameters' Comparison for different kernel sizes.}
\label{tab:table_6.0}    
\resizebox{0.7\columnwidth}{!} {
\begin{tabular}{llll}

\noalign{\smallskip}\hline\noalign{\smallskip} 
Layer                        & 3x3 filter & 5x5 filter & 7x7 filter \\ 
\noalign{\smallskip}\hline\noalign{\smallskip} 
Standard convolution         & 216        & 600        & 1176       \\ 
Involution                   & 145        & 289        & 505        \\ 
Depth aware hyper-involution & 273        & 273        & 273        \\ \hline
\end{tabular}
}   
\end{table*}

\subsection{Limitations $\text{\&}$  Future Directions}
The proposed model demonstrates strong and consistent performance across multiple RGB-D benchmarks, highlighting the effectiveness of integrating depth-aware hyper-involution into a single-stage detection framework. The experimental results indicate that the learned depth-guided spatial interactions provide meaningful complementary cues to RGB features, even when depth is estimated rather than directly sensed. In this study, depth maps are generated using DPT-hybrid, which offers a practical trade-off between accuracy and scalability in outdoor and mixed-scene settings. While pseudo-depth may exhibit local noise or irregularities, the proposed formulation leverages relative depth relationships rather than absolute depth values, contributing to its robustness across diverse scenes and datasets.

On SUN RGB-D, the method achieves competitive performance relative to recent RGB-D detectors. SUN RGB-D is characterized by substantial heterogeneity, as it combines depth measurements from multiple sensor types with different noise profiles and alignment characteristics. This variability particularly affects object categories with frequent occlusions, thin structures, or planar ambiguities. The observed performance trends suggest that the proposed depth-aware hyper-involution remains effective under such conditions, while methods employing heavier multi-stage cross-modal feature exchange may benefit from additional opportunities to smooth sensor-induced inconsistencies. These observations highlight complementary strengths among different RGB-D fusion strategies rather than fundamental limitations of the proposed approach.

The current architecture adopts a single fusion stage between RGB and depth features, balancing cross-modal interaction with computational efficiency. This design choice supports real-time inference while preserving a lightweight single-stage detection pipeline. Despite its simplicity, the proposed fusion mechanism consistently enhances detection performance over RGB-only baselines and introduces only marginal computational overhead compared to standard involution. While multi-level or iterative fusion schemes could further enrich cross-modal reasoning, the present results indicate that depth-aware hyper-involution already captures salient geometric cues in an efficient and scalable manner. Notably, fair comparison of model complexity and runtime across RGB-D detection methods remains challenging, as many recent works report results under non-unified hardware and evaluation protocols, underscoring the need for standardized benchmarking in future studies.

Finally, the proposed framework is inherently depth-model agnostic and can readily benefit from continued advances in monocular depth estimation. Recent depth models, such as Depth Anything V2 \cite{yang2024depth}, offer improved depth quality that may further strengthen the learned depth-aware interactions without requiring architectural changes. Beyond object detection, the proposed depth-aware hyper-involution provides a general mechanism for geometry-guided feature modulation and may be naturally extended to related tasks such as instance segmentation, part-level recognition, and structured scene understanding. These directions offer promising opportunities to further explore the versatility of the proposed design.

\section{Conclusions}

In this paper, we emphasized the significance of depth maps in object detection and explored convolution alternatives for improved feature extraction from RGB-D images. Introducing a depth-weighted hyper-involution and a novel fusion mechanism, we aimed to maximize depth information utilization, enabling dynamic learning and preventing information loss during model training. Leveraging these modules, we develop a lightweight single-stage RGB-D object detection model that ranks among the top-performing methods across benchmarks, while maintaining low computational complexity and a compact architecture. Qualitative and quantitative experiments on benchmark datasets validate the effectiveness of our architecture. Furthermore, we introduced two new RGB-D datasets, offering the research community more options for evaluation in diverse environments. How to further improve/adapt our depth-aware hyper-involution module for tasks such as object part segmentation or salient object detection is an interesting topic to explore. Future investigations could focus on its application in specific domains such as robotic surgery or augmented reality. In support of reproducible research and community use, we will publicly release the trained models together with the Outdoor \textit{RGB-D Detect} dataset.

\clearpage


\section*{Supplementary Information}
This supplementary material provides additional experimental details that complement the main manuscript. In particular, it describes a synthetic RGB-D data generation and evaluation framework designed to assess the generalization ability of the proposed method beyond standard indoor benchmarks. The supplement details the synthetic data generation pipeline, including CAD-based foreground projection, realistic scene composition, and depth map generation, it reports detection results on custom object categories. These experiments are intended to supplement the real-world outdoor RGB-D dataset presented in the main paper and provide further insights into the robustness of the proposed depth-aware detection framework in complex and industrial-style scenarios.

\subsection*{Synthetic Data Generation and Evaluation}

\begin{figure}
	\includegraphics[width=1.0\linewidth, height=0.4\linewidth]{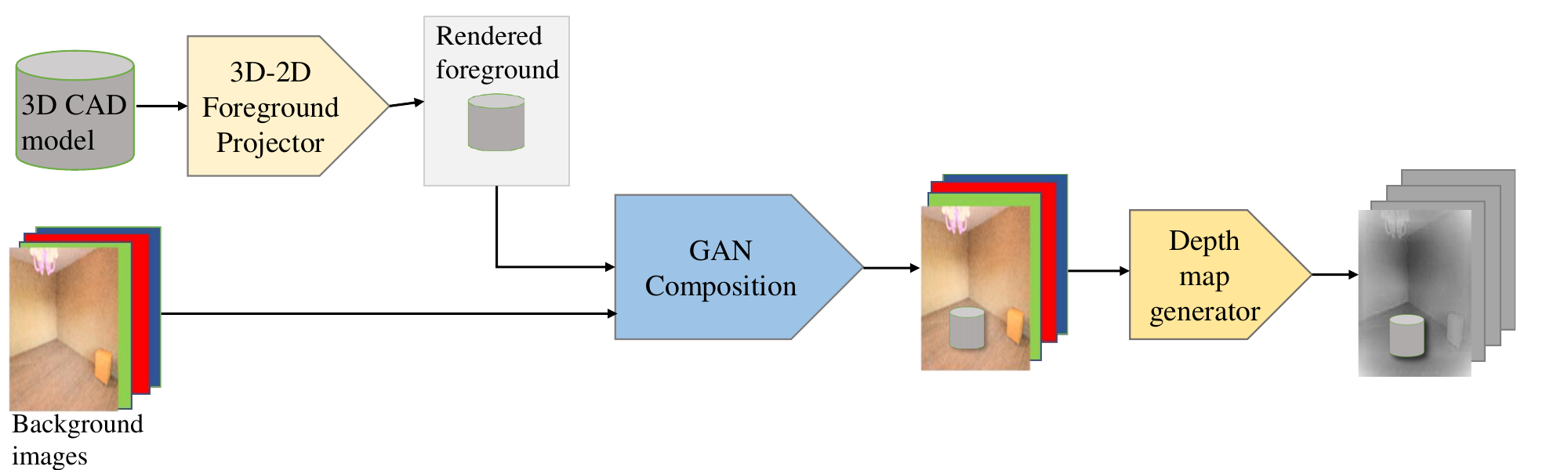}
	\caption[Automated RGB-D data generation pipeline]{Automated RGB-D data generation pipeline.}
	\label{DataGeneration}
\end{figure}

Previous RGB-D object detection methods primarily evaluated performance on benchmark datasets like SUN-RGBD and NYU Depth V2, which focus on indoor scenes with limited object types. This restricts the generalization of detectors to complex real-world scenarios with custom objects. Additionally, Industries often encounter object detection challenges due to a scarcity of relevant images. Thus, we devised a synthetic RGB-D data generation pipeline to assess our model's capability to detect custom objects in diverse environments. As demonstrated in Figure \ref{DataGeneration}, our RGB-D data generation framework consists of three main components. Firstly, a 3D-2D foreground projector for generating the perspective projections of 3D CAD (Computer Aided Design) models. Then, a generative composition model to create realistic composite images of the projected foreground image with selected background images. The depth map generator produces depth maps for composite images. Using a 3D CAD model input, the 3D-2D foreground projector module generates 2D viewpoint images, utilizing parameters like azimuth, elevation, and distance. It also incorporates six degrees of freedom to capture various orientations of the CAD models when generating their silhouettes. Next, we apply the Spatial Transformer Generative Adversarial Network (ST-GAN) \cite{lin2018st} to combine our generated foreground image with the background image while maintaining the geometric correction and the scene semantics. Then we utilize a hybrid version of the dense vision transformer (DPT-hybrid) \cite{ranftl2021vision} as our final component, i.e., the depth map generator. DPT-hybrid transforms composite RGB images into tokens using ResNet-50 \cite{he2016deep} to generate aligned depth maps efficiently. The pipeline generated 16,000 RGB-D data in just 3 minutes on an Nvidia Quadro RTX 6000 GPU, showcasing its industry utility.

\begin{table*}[h]	
	\center
	\resizebox{0.9\linewidth}{!}
	{
		\begin{tabular}{|p{1.79cm}|p{0.9cm}|p{2.1cm}|p{0.8cm}|p{1.2cm}|p{1.9cm}|p{0.8cm}|p{1.8cm}|p{0.55cm}|}
			\hline
			Method & \textbf{mAP} & doorstopper & pipe & clamp & screwdriver & brace & paintbrush & nut \\
			\hline
			\hline
			FETNet \cite{xiao2021fetnet} & 56.8 & 80.6 & 71.3 & 59.6 & 68.1 & 49.6 & 54.7 & 14.3 \\
			\hline
			\textbf{Ours} & 58.9 & 84.1 & 74.9 & 67.2 & 62.7 & 53.0 & 52.5 & 17.9 \\
			\hline
		\end{tabular}
	}
	\caption{\label{tab:table_4.1} Experimental results on automatically synthesized dataset.}
\end{table*}
\subsection*{Results on Synthetic Dataset}

\begin{figure}
	\centering
	\includegraphics[width=8.75 cm]{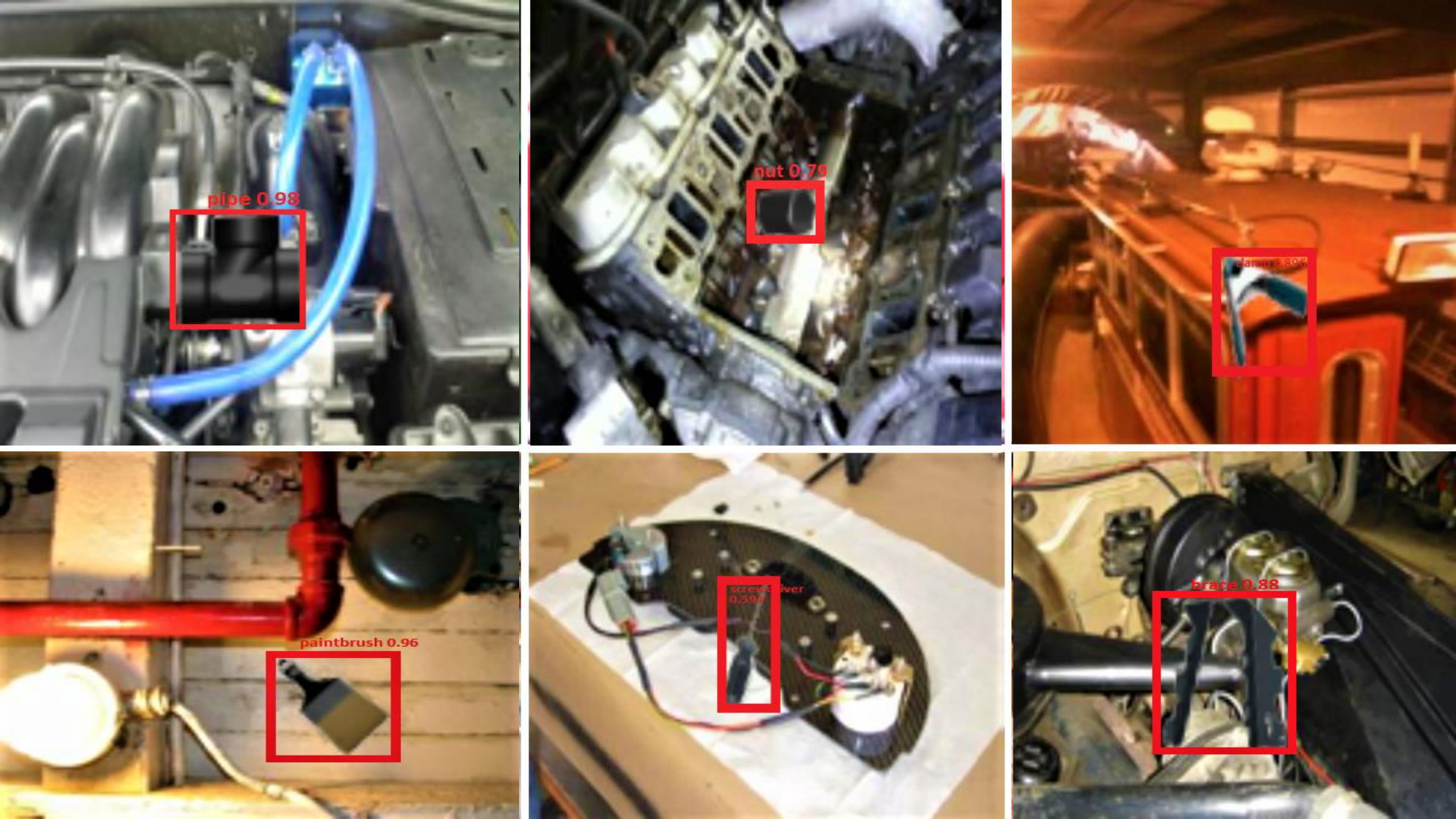}
	\caption[Detection results on the synthesized dataset.]{A few detection results on the synthesized data.}
	\label{Detection_result2}
\end{figure}

Our model achieves an overall mAP of 58.7$\%$ on seven small working object classes (clamp, pipe, brace, nut, screwdriver, door-stopper, and paintbrush) from synthesized data, as shown in Table \ref{tab:table_4.1}. Figure \ref{Detection_result2} shows qualitative detection results, showing red boxes on synthetic data. While mAP is lower for very small objects like nuts due to depth data noise, notably, the model performs well for individual small object classes like doorstopper, brace, and clamp in complex synthetic factory environments.

\backmatter
\bmhead{Acknowledgements}

This work was supported by The Mathematics of Information Technology and Complex Systems (MITACS) through the Mitacs Accelerate program \#IT21451.

\bibliography{egbib}

\end{document}